\documentclass[journal]{IEEEtran}

\ifCLASSINFOpdf
\else
\fi

\usepackage{amssymb}

\usepackage{amsmath}
\usepackage{amssymb}
\usepackage{amstext}
\usepackage{amsfonts}
\usepackage{graphicx}
\usepackage{epsfig}
\usepackage{subfig}
\usepackage{xcolor}
\usepackage{rotating}
\usepackage{soul}
\usepackage{upgreek }
\usepackage{setspace}
\usepackage[switch,columnwise]{lineno}

\usepackage{hyperref,graphicx,enumitem}

\newcommand{\norm}[1]{\left\lVert#1\right\rVert}

\begin{document}

\title{A Multisensor Hyperspectral Benchmark Dataset For Unmixing of Intimate Mixtures}

\author{Bikram~Koirala,
        Behnood~Rasti,
        Zakaria~Bnoulkacem,
        Andréa~de~Lima~Ribeiro,
        Yuleika~Madriz,
        Erik~Herrmann,
        Arthur~Gestels,
        Thomas~De~Kerf,
        Sandra~Lorenz,
        Margret~Fuchs,        
        Koen~Janssens,
        Gunther~Steenackers,
        Richard~Gloaguen,        
        and Paul~Scheunders        
\thanks{Bikram Koirala (corresponding author), Zakaria Bnoulkacem, and Paul Scheunders are with Imec-Visionlab, University of Antwerp (CDE), Universiteitsplein 1, B-2610 Antwerp, Belgium; Bikram.Koirala@uantwerpen.be; zakaria.bnoulkacem@uantwerpen.be; paul.scheunders@uantwerpen.be}
 \thanks{Behnood Rasti, Andréa de Lima Ribeiro, Yuleika Madriz, Erik Herrmann, and Richard Gloaguen    are with Helmholtz-Zentrum Dresden-Rossendorf, Helmholtz Institute Freiberg for Resource Technology, Machine Learning Group, Chemnitzer Straße 40, 09599 Freiberg, Germany; b.rasti@hzdr.de; a.de-lima-ribeiro@hzdr.de; y.madriz-diaz@hzdr.de; e.herrmann@hzdr.de; 
s.lorenz@hzdr.de; m.fuchs@hzdr.de; r.gloaguen@hzdr.de}
\thanks{Arthur Gestels and Koen Janssens are with Antwerp X-ray Imaging and Spectroscopy (AXIS), University of Antwerp, Groenenborgerlaan 171, B-2020 Antwerp, Belgium;   arthur.gestels@uantwerpen.be; koen.janssens@uantwerpen.be}
\thanks{Thomas De Kerf and Gunther Steenackers are with InViLab research group, University of Antwerp, Groenenborgerlaan 171, B-2020 Antwerp, Belgium;   thomas.dekerf@uantwerpen.be; gunther.steenackers@uantwerpen.be}
}

\maketitle

\begin{abstract}
Optical hyperspectral cameras capture the spectral reflectance of materials. Since many materials behave as heterogeneous intimate mixtures with which each photon interacts differently, the relationship between spectral reflectance and material composition is very complex. Quantitative validation of spectral unmixing algorithms requires high-quality ground truth fractional abundance data, which are very difficult to obtain. 

In this work, we generated a comprehensive laboratory ground truth dataset of intimately mixed mineral powders. For this, five clay powders (Kaolin, Roof clay, Red clay, mixed clay, and Calcium hydroxide) were mixed homogeneously to prepare 325 samples of 60 binary, 150 ternary, 100 quaternary, and 15 quinary mixtures. Thirteen different hyperspectral sensors have been used to acquire the reflectance spectra of these mixtures in the visible, near, short, mid, and long-wavelength infrared regions (350-15385) nm. Overlaps in wavelength regions due to the operational ranges of each sensor and variations in acquisition conditions resulted in a large amount of spectral variability. Ground truth composition is given by construction, but to verify that the generated samples are sufficiently homogeneous, XRD and XRF elemental analysis is performed. We believe these data will be beneficial for validating advanced methods for nonlinear unmixing and material composition estimation, including studying spectral variability and training supervised unmixing approaches. The datasets can be downloaded from the following link: \url{https://github.com/VisionlabUA/Multisensor_datasets}.
\end{abstract}

\begin{IEEEkeywords}
Hyperspectral, intimate mixtures, multi-sensor dataset, benchmark, unmixing
\end{IEEEkeywords}

\IEEEpeerreviewmaketitle

\section{Introduction}
\label{sec:intro}
\IEEEPARstart{S}{ince} each material interacts differently with incident light, it can be uniquely characterized by its reflectance spectrum. In remote sensing, hyperspectral cameras (HSC) are used to resolve the reflected sunlight into hundreds of successive small wavelength bands in the visible and near-infrared (VNIR, (350-1000) nm) and the short-wave infrared (SWIR, (1000-2500) nm) regions \cite{Gham17}. Due to limitations of the sensor's spatial resolution, a pixel may contain more than one material. The measured spectral reflectance is generally modeled as a linear mixture of the different materials involved \cite{Biou12}. The linear mixing model has shown very good performance in scenarios where the Earth's surface contains large homogeneous areas with clearly separated regions containing different pure materials \cite{RHeylen2016}. In close-range scenarios, however, the material behaves as an "intimate mixture" with which each photon interacts differently, making the relationship between the spectral reflectance and material composition highly complex and highly nonlinear \cite{abiliointimate}.

With advances in technology, small portable, low-cost HSC have emerged that can be installed on unmanned aerial vehicles, agricultural machinery, conveyors, or used in laboratory environments \cite{Kurz16}. The close-range setting generates higher-quality reflectance spectra, better describing the intimate mixtures, and can potentially estimate material composition from spectral reflectance \cite{KRUPNIK2019102952}.

Several nonlinear unmixing approaches have been developed \cite{Hey14}. Nonlinear mixing models, describing secondary reflections \cite{Dob14}, or higher-order interactions simplify the complex interaction of light with mixtures \cite{Hey19, AMarinoni20169}. The study \cite{HapkeCNN} introduced the HapkeCNN, a Convolutional Neural Network (CNN) that incorporates the Hapke model into the learning process. By integrating the physical model into the CNN architecture, the HapkeCNN enhances the understanding and representation of the underlying physical processes involved in the data. Although physics-based mixing models have been developed to explain the interaction of light with intimate mixtures \cite{Hapke81, Hapke98}, most of them fail when the material grains/particles have a size, smaller or similar to the wavelength of light, are non-spherically shaped and behaving as anisotropic scatterers. Moreover, these nonlinear mixing models are not invariant to spectral variability caused by changes in acquisition conditions, such as illumination conditions, distance, and orientation from the sensor, using different sensors or white calibration panels. {\color{black} In this work, we will use the term external variability to refer to spectral variability caused by changes in acquisition conditions.}

In previous work, we developed a robust nonlinear unmixing approach by combining model-based unmixing and machine learning and validated it on binary mineral mixtures \cite{Koi21}. Accurate characterization of mixtures of more than two pure materials is currently lacking in the literature. 

To develop and validate advanced algorithms that can characterize mixtures of more than two pure materials and also address external variability, a comprehensive hyperspectral dataset of intimate mineral powder mixtures acquired by multiple sensors with different acquisition configurations is required. 

Recently, research has been devoted to generating ground-truth data for spectral mixture analysis in the remote sensing field; this, however, remained limited to linearly mixed data. In \cite{Kumar22}, targeted scenes containing paper-based panels of different sizes and filled with different colors and proportions were imaged with a terrestrial hyperspectral imager at various spatial resolutions.
In \cite{DCerra21}, a hyperspectral unmixing dataset was generated by data acquired over the German Aerospace Center premises in Oberpfaffenhofen, containing airborne hyperspectral images, high-resolution RGB images, and ground-measured spectral reflectance of a number of synthetic reference targets of different materials and sizes.

Several recent works report on the generation of small-scale intimately mixed datasets. The Relab data set contains spectra of crafted mineral mixtures from the NASA Reflectance Experiment Laboratory (RELAB) at Brown University, publicly available at www.planetary.brown.edu/relab/ \cite{Mustard1989PhotometricPF}. 
The data set contains binary mixtures from five minerals: Anorthite (An), Bronzite (Br), Olivine (Ol), Quarts (Qz)
and Alunite (Al). For each of the binary mixtures of An-Br, Br-Ol, Ol-An, and Qz-Al, 3 mixtures were
available with a 25\%, 50\%, and 75\% ratio by mass.
{\color{black} In \cite{EDucasse2020}, a spectral database was built by acquiring hyperspectral images from 29 intimate binary and ternary mixtures by homogeneously mixing five different clay samples (illite, kaolinite, montmorillonite, calcite, and quartz). }
{\color{black} In \cite{MinZhao19}, different mixing scenes were set up to mimic various mixture models. Hyperspectral images were acquired from a checkerboard (linear mixtures), a scene containing a vertical board and the checkerboard (bilinear mixtures), and mixed quartz sands (intimate mixtures). Endmembers of these scenes were collected by acquiring images of pure materials. The ground truth fractional abundances of the intimate mixtures were obtained from high-resolution RGB images. For the other scenes, ground truth fractional abundances were given on the basis of the material composition.} In \cite{Koi21}, we developed datasets of homogeneously mixed mineral powder mixtures acquired by two sensors, i.e., an AgriSpec spectroradiometer [manufactured by Analytical Spectral Devices (ASD)] and a snapscan shortwave infrared hyperspectral camera, under strictly controlled experimental settings. The data set contained a total of 49 binary mixtures of 5 mineral powders. The five chosen minerals are different oxides, typically found in soil and applied in cementitious materials: Aluminum oxide (Al$_2$O$_3$), Calcium oxide (CaO), Iron oxide (Fe$_2$O$_3$), Silicon dioxide (SiO$_2$) and Titanium dioxide (TiO$_2$). In total, seven binary mixture combinations of minerals were prepared: Al$_2$O$_3$-SiO$_2$ (Al-Si), CaO-SiO$_2$ (Ca-Si), CaO-TiO$_2$ (Ca-Ti), Fe$_2$O$_3$-Al$_2$O$_3$ (Fe-Al), Fe$_2$O$_3$-CaO (Fe-Ca), Fe$_2$O$_3$-SiO$_2$ (Fe-Si) and SiO$_2$-TiO$_2$ (Si-Ti). For each mineral combination, 7 different mixtures were prepared.

\subsection{Contributions and Novelties}
{\color{black} The contribution of this work is fourfold.
\begin{enumerate}
    \item Comprehensive intimate mixture samples: In this work, we generated a comprehensive hyperspectral dataset of intimate mineral powder mixtures by homogeneously mixing five different clay powders (Kaolin, Roof clay, Red clay, mixed clay, and Calcium hydroxide) in laboratory settings. In total 325 samples were prepared. Among them, 60 mixtures are binary, 150 mixtures are ternary, 100 mixtures are quaternary, and 15 mixtures are quinary. To the best of our knowledge, this is the first publicly available intimate mixture dataset of this size, generated by mixing up to five pure materials. {\color{black} Because these pure clay powders are typically applied in building materials, the quality of construction materials is determined by the amount (fractional abundances) of pure clay powder in the mixture. 
    }
    \item Comprehensive sensor measurements: For each mixture (and pure clay powder), reflectance spectra are acquired by 13 different sensors, with a broad wavelength range between the visible and the long-wavelength infrared regions (i.e., between 350 nm and 15385 nm) and with a large variation in sensor types, platforms, and acquisition conditions. We believe that this dataset will help researchers to develop single and multi-sensor hyperspectral unmixing algorithms that deal with the large spectral variability and intrinsic nonlinearities of the data. {\color{black} We like to clarify that we do not target typical remote sensing applications but rather applications in close-range scenarios. As each sample is a homogeneous mixture of different pure clay powders, the spatial scope of these datasets is limited. When acquired by imagers, the mean spectrum of each sample can be used to analyze the mixture. 
    }    
    
    \item{{\color{black} The dataset of intimate mixtures challenges existing unmixing methodologies, as demonstrated by showing highly inaccurate results from linear and Hapke-based unmixing.}}
    
    \item Confirmed ground truth fractional abundances: High precision ground truth of the fractional abundances is given by construction. {\color{black} To validate the ground truth for the elemental composition of each powder}, X-ray powder diffraction (XRD) and X-ray fluorescence (XRF) elemental analysis are performed on all mixtures. 

\end{enumerate}
}

This manuscript is organized as follows: section \ref{Sample preparation} presents a detailed description of the sample preparation; section \ref{VNIR and SWIR sensors} describes the data acquisition in the VNIR and SWIR wavelength regions; section \ref{MWIR and LWIR sensors} describes the data acquisition in the mid-wave infrared (MWIR, (2500-8000) nm) and long-wave infrared (LWIR, (8000-15000) nm) wavelength regions; section \ref{X-ray sensors} describes the data acquisition with the X-ray sensors.

\section{Sample preparation}
\label{Sample preparation}

In this work, we used 5 pure clays typically applied in building materials: Kaolin (Ka), Roof clay (Ro), Red clay (Re), mixed clay (Mi), and Calcium hydroxide (Ca). Kaolin is a mixture of Aluminium silicate hydroxide and Silicon dioxide while Roof clay, Red clay, and mixed clay mostly contain Aluminium silicate hydroxide, Biotite, Goethite, and Silicon dioxide. We sieved these  clay samples with a sieve with 200 $\mu$m openings, hereby limiting the grain size of all samples to less than 200 $\mu$m. We estimated the particle density of each pure clay in the lab with a pycnometer, following the standard soil particle density protocol \cite{hao2008soil}.

{\color{black} We generated a total of 325 mixtures by mixing these five pure clay powders.} All possible clay combinations of these powders were considered, i.e., 10 binary combinations, 10 ternary combinations, 5 quaternary combinations, and one quinary combination (see Table \ref{Mixture summary}).

	\begin{table}
		\caption{{\color{black} Summary of the clay mixtures of Kaolin (Ka), Roof clay (Ro), Red clay (Re), mixed clay (Mi), and Calcium hydroxide (Ca)}}
		\centering
		\scalebox{0.9}{
		\begin{tabular}{|c|c|}
			\hline
			Mixture type      			            & Clay combinations\\   
			\hline
   		Binary  & 1) Ka-Ro; 2)  Ka-Re; 3)  Ka-Mi;\\
                    & 4)  Ka-Ca; 5)  Ro-Re; 6) Ro-Mi;\\
                    & 7)  Ro-Ca; 8)  Re-Mi; 9)  Re-Ca;\\
                    & 10)  Mi-Ca\\
                \hline
		    Ternary  & 1)  Ka-Ro-Re; 2) Ka-Ro-Mi;\\
                     & 3)  Ka-Ro-Ca; 4) Ka-Re-Mi;\\
                     & 5) Ka-Re-Ca; 6) Ka-Mi-Ca;\\
                     & 7)  Ro-Re-Mi; 8)  Ro-Re-Ca;\\
                     & 9)  Ro-Mi-Ca; 10)  Re-Mi-Ca\\
			\hline
			Quaternary  & 1)  Ka-Ro-Re-Mi; 2)  Ka-Ro-Re-Ca; \\
                        & 3)  Ka-Ro-Mi-Ca; 4) Ka-Re-Mi-Ca;\\
                        & 5)  Ro-Re-Mi-Ca\\
			\hline
			Quinary  & 1)  Ka-Ro-Re-Mi-Ca \\
			\hline			
		\end{tabular}    }
		\label{Mixture summary}
	\end{table}
	

Within each clay combination, samples with different mixture fractions are generated so that the ground truth fractional abundances uniformly cover the five-dimensional simplex, with a step size of 14.286 \% mass ratios. In this way, 6 unique mixtures are generated for each binary clay combination, 15 for each ternary clay combination, 20 for each quaternary clay combination, and 15 for the combination of 5 clays, making a total of 325 mixtures.

\begin{figure}[htb!]
\centering
\begin{tabular}{l}
	\includegraphics[width=.45\textwidth]{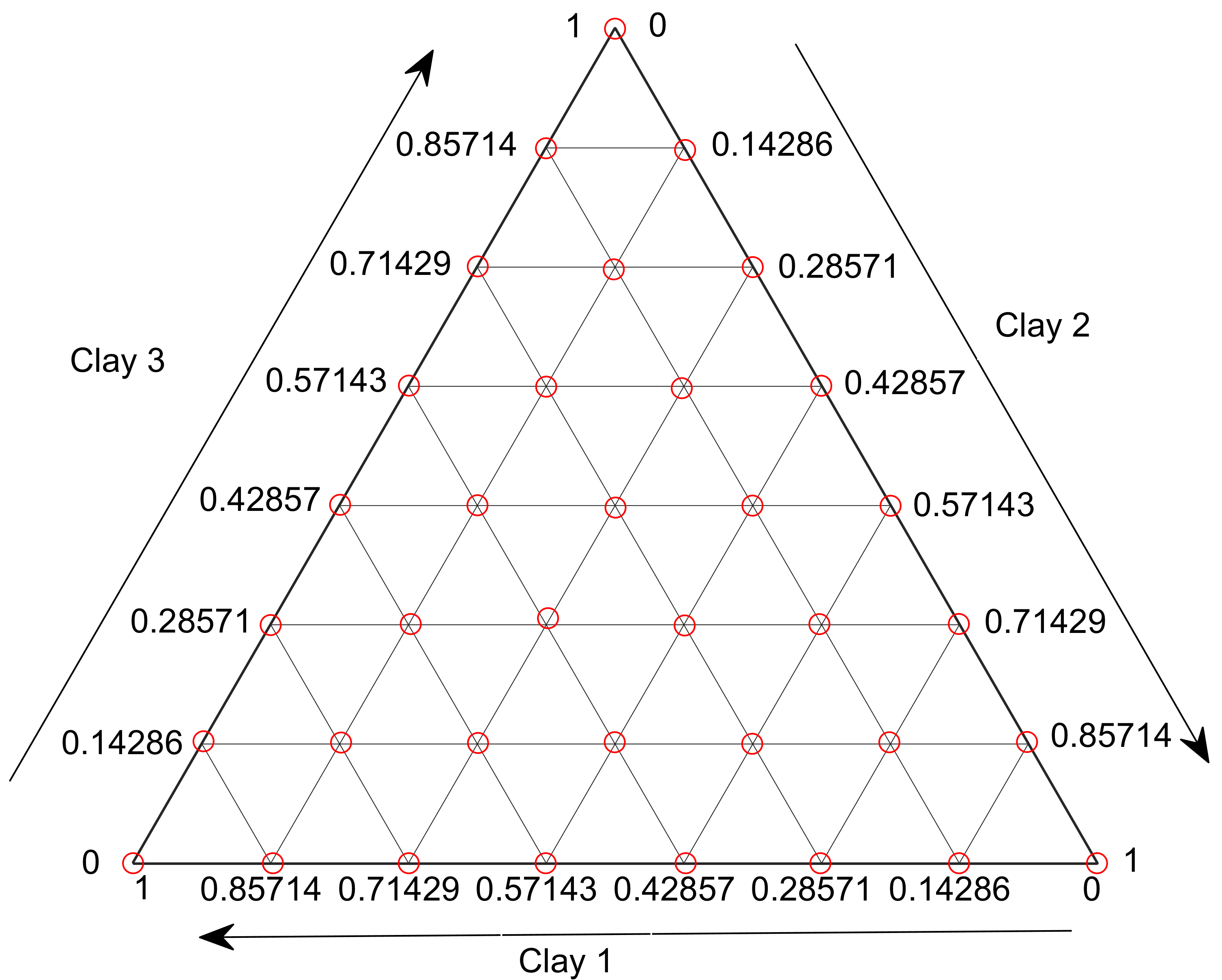}\\
 \end{tabular}
\caption{The ternary diagram of clay mixtures.}
\label{Ternary simplex}
\end{figure}

In Fig. \ref{Ternary simplex} we display the uniformly sampled fractional abundances for a ternary clay combination. The three clays occupy the corners of the simplex, all binary mixtures lie on the lines connecting two clay's while ternary mixtures lie inside the simplex. 

Mixtures were produced by weighing and aggregating the different clays. We fixed the weight of each mixture sample to be a total of 10 g, the scale had a precision of 0.001 g.  Each 10 g sample was placed in a glass bottle and a homogeneous mixture was produced by rotating the bottle for approximately five minutes. 

Using the particle densities of the pure clays, we converted the weight to volume fraction by:
\begin{equation}
a_{j} = \frac{\frac{M_j}{\rho_j}}{\sum_{j=1}^{p} \frac{M_j}{\rho_j}},
\label{Volume fraction}
\end{equation}
where $p$ denotes number of pure materials, $M_j$ is the mass fraction of component $j$, and $\rho_j$ is its density.

\begin{figure}[htb!]
\centering
\begin{tabular}{l}
	\includegraphics[width=.45\textwidth]{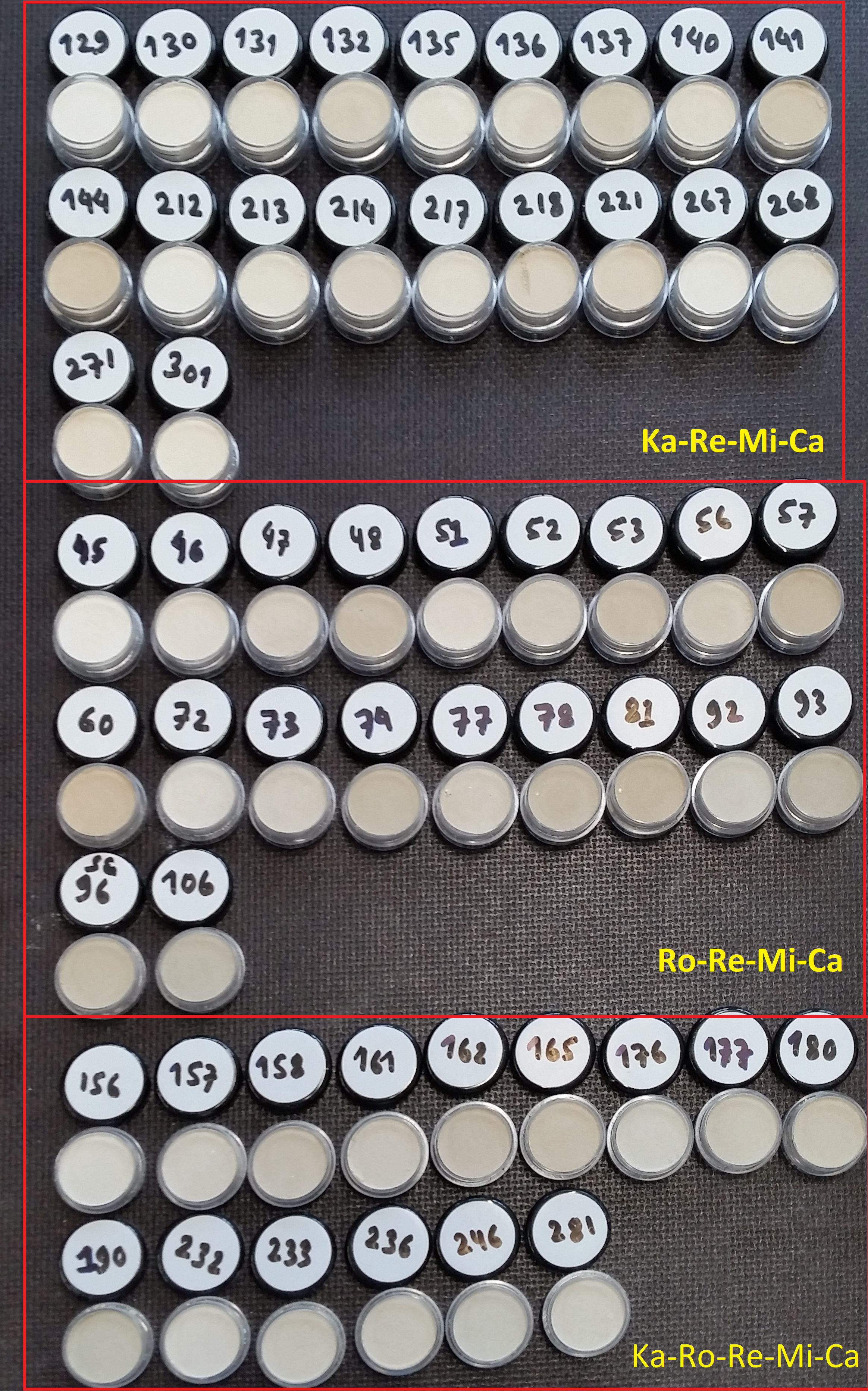}\\
 \end{tabular}
\caption{{\color{black} The RGB image of 40 quaternary mixtures (Ka-Re-Mi-Ca and Ro-Re-Mi-Ca) and all 15 quinary mixtures.} Ka, Ro, Re, Mi, and Ca refer to Kaolin, Roof clay, Red clay, Mixed clay, and Calcium hydroxide respectively.}
\label{RGB_image_sample}
\end{figure}

Each sample was then placed in a clear plastic sample holder with an inner diameter of 3.048 cm and a height of 1.524 cm. Approximately 3 g of a mixture was required to fill the sample holder. The samples were then compacted and smoothed using a stamp compactor. In Fig.  \ref{RGB_image_sample}, the 40 samples of 2 quaternary clay combinations (Ka-Re-Mi-Ca and Ro-Re-Mi-Ca) and all 15 samples of the quinary combination are shown. 

The 325 mixtures and the 5 pure clay powders were scanned using 13 different sensors covering a broad range between the visible and the LWIR wavelength regions (350 nm to 15385 nm). For all samples, the ground truth composition is given by construction, but to verify that the generated samples are sufficiently homogeneous,  X-ray powder diffraction and X-ray fluorescence elemental analysis (Bruker Tornado M4) were performed. Table \ref{Sensor summary} summarizes the properties of all the sensors.
	\begin{table*}
		\caption{{\color{black} Summary of sensors employed in this investigation}}
		\centering
		\scalebox{0.9}{
		\begin{tabular}{|c|c|c|c|c|}
			\hline
			Sensor      			            & Spectral range & Bands/Channels   & Spatial resolution  & Spectral resolution\\   
			\hline
   			ASD Spectroradiometer  & 350 nm to 2500 nm  & 2151  & - & 3-6 nm \\
                \hline
		    Senops HSC2  & 500 nm to 900 nm  & 50  & 1024 $\times$ 1024 pixels & 10-16 nm \\
			\hline
			Specim sCMOS  & 400 nm to 1000 nm  & 238  & 2148 pixels & 6 nm \\
			\hline
			Specim AisaFenix  & 400 nm to 2500 nm  & 450  & 1024 pixels & 3.5-10 nm \\
			\hline
			Specim FX50  & 2700 nm to 5300 nm  & 308  & 640 pixels & 35 nm \\
			\hline
			Specim AisaOwl  & 7600 nm to 12300 nm  & 96  & 385 pixels & 100 nm \\
			\hline
			Telops MWIR  & 3000 nm to 9000 nm  & 110  & 320 pixels & 20 cm$^-1$ \\
			\hline
			Telops HyperCam  & 7400 nm to 12500 nm  & 128  & 320 $\times$ 256 pixels & 2 cm$^-1$ \\
			\hline
   			Specim JAI(RGB)   & 440 nm to 630 nm  & 3  & 4096 $\times$ 8496 pixels & - \\
			\hline
   			PSR-3500 spectral evolution  & 350 nm to 2500 nm  & 1024  & - & 2.8-8 nm \\
			\hline
   			Agilent 4300 FTIR  & 2500 nm to 15385 nm  & 7191  & - & 4-16 cm$^-1$ \\
			\hline
   			Cubert Ultris X20P  & 350 nm to 1000 nm  & 164  & 410 $\times$ 410 pixels & 4 nm \\
			\hline
   			Cubert panchromatic  & 350 nm to 1000 nm & 1  & 1886 $\times$ 1886 pixels & - \\
			\hline
   			micro-X-ray fluorescence  & -  & -  & - & - \\
                spectrometer (Tornado)                  &    &    &   &   \\
			\hline
   			Macro X-ray powder diffractometer  & -  & -  & - & - \\
			\hline
		\end{tabular}    }
		\label{Sensor summary}
	\end{table*}

\section{Data acquisition in VNIR and SWIR wavelength regions}
\label{VNIR and SWIR sensors}
\subsection{Handheld spectroradiometers}
We used two handheld spectroradiometers (ASD and PSR-3500 spectral evolution) to acquire the spectral reflectance of our 330 samples in the VNIR and SWIR wavelength regions. We acquired the spectra in indoor environments, and the illumination source for the ASD spectroradiometer is the ASD Muglight. The PSR-3500 spectral evolution sensor contains an illumination source attached to the sensor itself. Since these sensors acquire radiance spectra, the spectral reflectance is obtained according to the equation \eqref{Radiance}:
\begin{equation}
\mathbf{y} = \frac{\mathbf{R}_{sample} - \mathbf{R}_{dark}}{\mathbf{R}_{white} - \mathbf{R}_{dark}} \times \mathbf{y}_{spec},
\label{Radiance}
\end{equation}
where $\mathbf{y}$ is the calibrated spectral reflectance, $\mathbf{R}_{sample}$ is the radiance of the sample, $\mathbf{R}_{white}$ is the radiance of a white calibration panel, i.e., a highly reflecting surface (e.g., Spectralon), $\mathbf{y}_{spec}$ is the reflectance of the Spectralon, and $\mathbf{R}_{dark}$ is the radiance, acquired by the sensor when the light source is turned off. 

Since spectral reflectances of most of the natural materials are dependent on the illumination and acquisition angles, for both handheld sensors, the data is acquired approximately orthogonal to the sample surface. The illumination angle is kept constant for each sensor and was 35$^0$ and 0$^0$ for the ASD spectroradiometer and the PSR-3500 spectral evolution, respectively.

The spectral reflectance will also vary with the distance of the sample to the sensor, described by the following equation:
\begin{equation}
\mathbf{y}_{meas} = \frac{\mathbf{y}}{(1 + \frac{d_{sample}-d_{ref}}{d_{ref}})^2} ,
\label{Scaling}
\end{equation}
where $\mathbf{y}_{meas}$ is the measured spectral reflectance, $d_{ref}$ is the distance between the white calibration panel and the sensor, and $d_{sample}$ is the distance between the sample and the sensor.
The distance variation is reflected in a scaling of the spectrum (see equation \ref{Scaling}). {\color{black} Samples heights varied due to two main reasons}:
variations in the compaction of each sample and the difference in clay densities.
Since the distance between the sample and the sensors in the handheld devices is of the order of a few centimeters, a variation of the height of the sample in the order of a few millimeters can cause a significant scaling effect. {\color{black} Our investigations did not include measurements of the sample's heights, hence the random scaling effect must be regarded as external spectral variability.}

\subsection{Imaging sensors}
\begin{figure*}[htbp]
\centering
 \begin{tabular}{c}
 	\includegraphics[width=.9\textwidth]{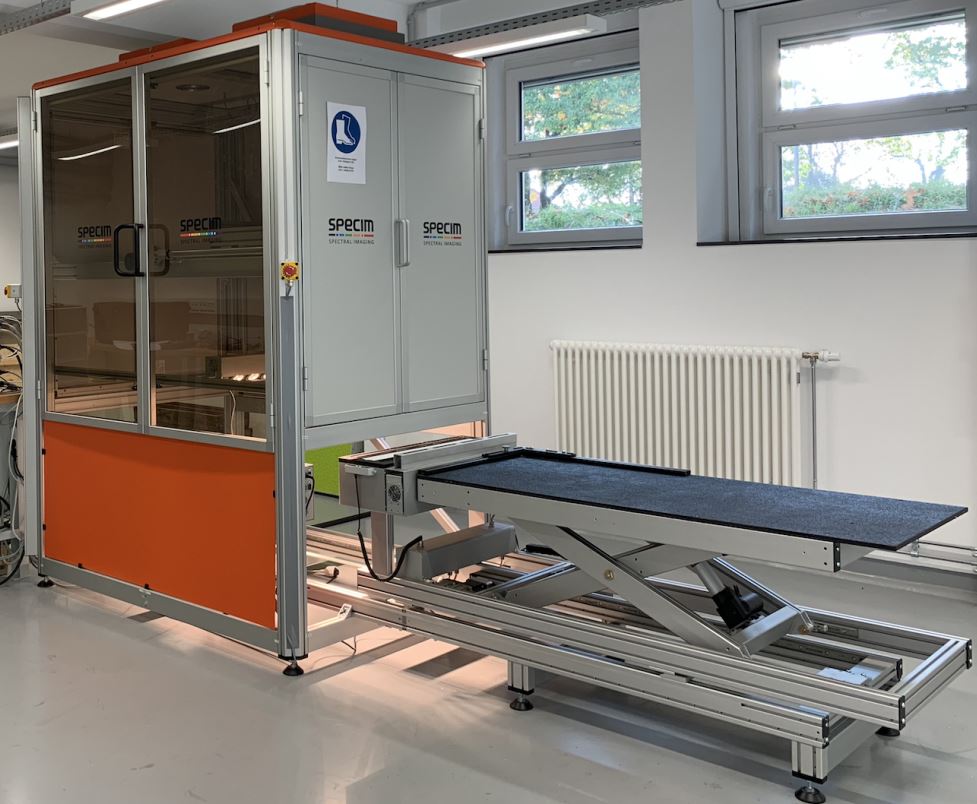} \\ 
 (a) \\
 \includegraphics[width=.9\textwidth]{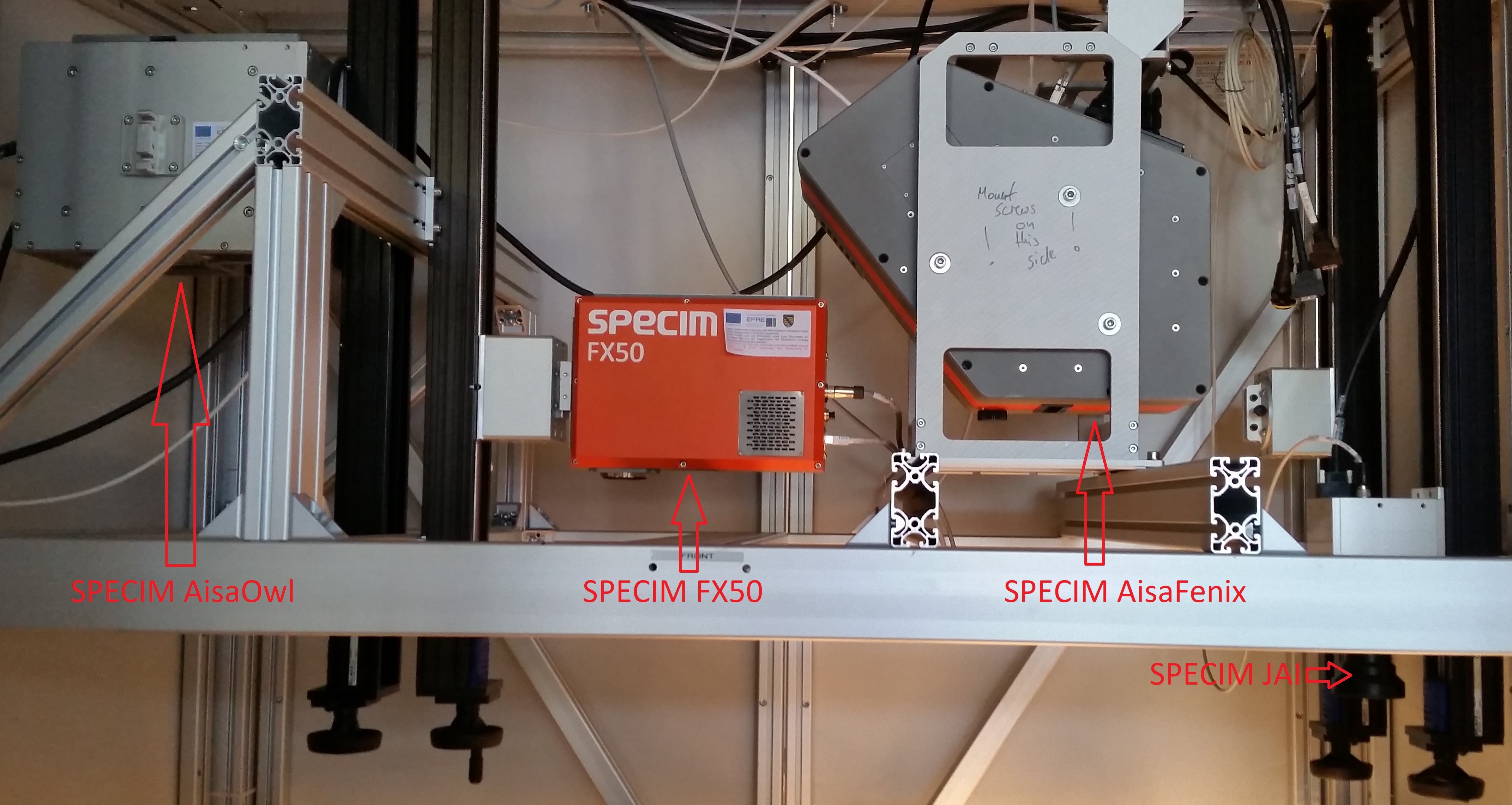}\\
 (b)
  \end{tabular}
 \caption{Drill-core scanner equipped with multiple sensors. (a) Entire drill-core scanner platform; (b) Drill-core scanner equipped with four different sensors (Specim JAI, Specim AisaFenix, Specim FX50, and Specim AisaOwl). }
\label{Drill core scanner}
\end{figure*}

Unlike handheld sensors that can only collect spectra from one sample at a time, imaging sensors can collect images from multiple samples simultaneously. {\color{black} The spectral reflectances were acquired from the samples by 6 different imaging sensors}, using two different scanning setups, developed at the Helmholtz Institute Freiberg for Resource Technology. The first scanner is a drill core scanner (see figure \ref{Drill core scanner}), in which a Specim AisaFenix hyperspectral camera and Specim JAI RGB camera were mounted. This scanner carries samples on a moving table under the field of view of these two cameras. The distance between the Specim AisaFenix and the moving table is approximately 0.93 m, while the Specim JAI RGB is approximately 0.63 m away from the moving table. The Specim AisaFenix hyperspectral camera contains two different sensors that cover the VNIR and SWIR wavelength regions, respectively. The spectral resolution of the sensors is approximately 3.5 nm in the VNIR and 12 nm in the SWIR. The ground sampling distance (GSD) of each hyperspectral pixel is approximately 1.5 mm while each RGB pixel covers approximately 0.1 mm on the ground.

\begin{figure}[htb!]
\centering
 \begin{tabular}{l}
 	\includegraphics[width=.48\textwidth]{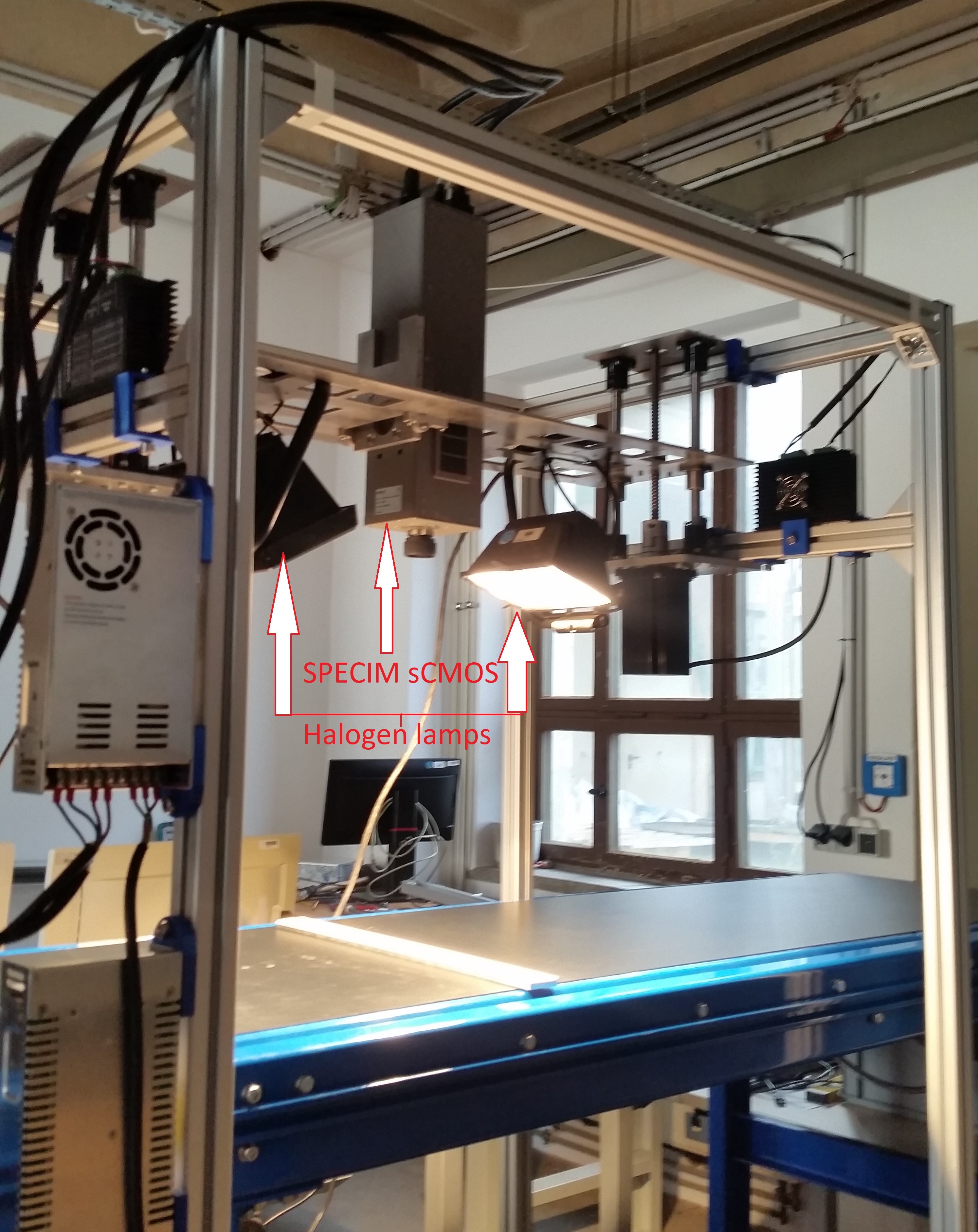} 
  \end{tabular}
 \caption{Conveyor belt scanner with the Specim sCMOS camera mounted on it.}
\label{Conveyor belt}
\end{figure}

The second scanner is a conveyor belt scanner (see Fig. \ref{Conveyor belt}), on which the following four sensors can be mounted: Specim sCMOS, Cubert Ultris X20P, Cubert panchromatic, and Senops HSC2. Only one sensor at a time can be mounted on the scanner. Fig. \ref{Conveyor belt} shows the Specim sCMOS. The distance between the Specim sCMOS and the moving belt is approximately 0.45 m. The characteristics of these four sensors are summarized in Table \ref{Sensor summary}. This platform brings samples toward the camera by moving the belt. Two halogen lamps were installed to provide an artificial light source. 

We placed our samples on wooden plates to minimize the motion effects during the acquisition of the reflectance data (see Fig. \ref{RGB_image_sample} as an example). Only three wooden plates were needed to fit all 330 samples. After acquiring the images, a rectangular bounding box with a limited number of pixels was selected from the center of each sample. This procedure is essential to remove unrelated objects (edges of the sample holders) and shadow areas from the image. Due to variations in the spatial resolution of the cameras, the final image size of each sample ranges from 8 $\times$ 8 pixels to 56 $\times$ 66 pixels. The radiance images acquired by these cameras have been converted to reflectance using an internal workflow based on Hylite \cite{THIELE2021104252}. Spectra are averaged over the entire sample image.

\subsection{Spectral reflectance of pure clay samples}
\begin{figure*}[htbp]
\centering
 \begin{tabular}{ccc}
 	\includegraphics[width=.32\textwidth]{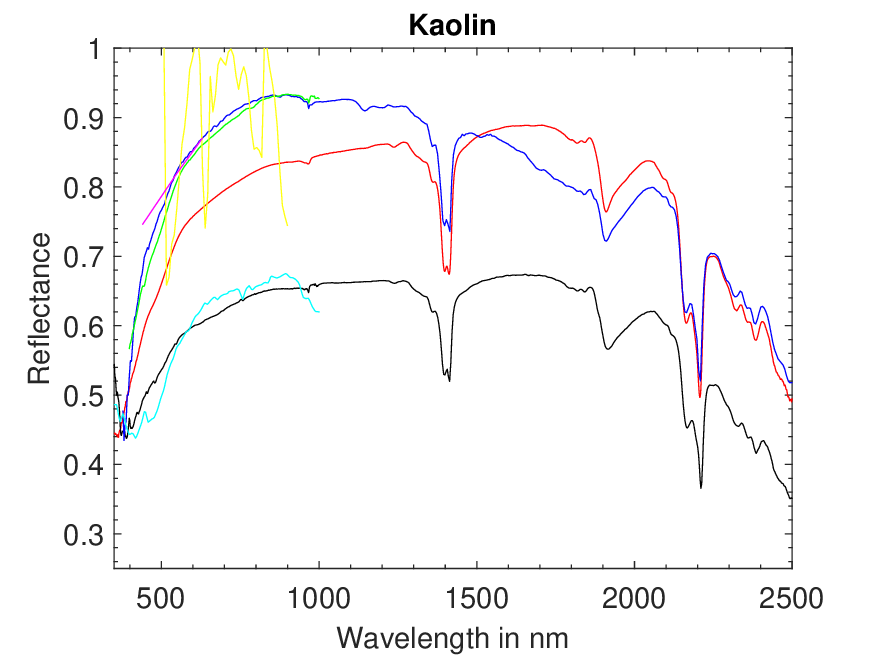} & 	\includegraphics[width=.32\textwidth]{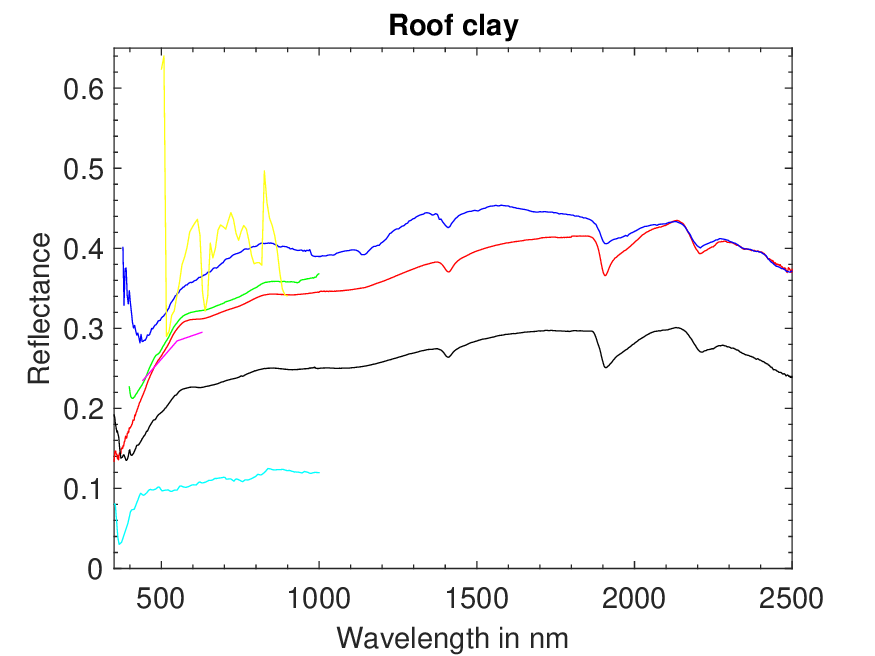} & \includegraphics[width=.32\textwidth]{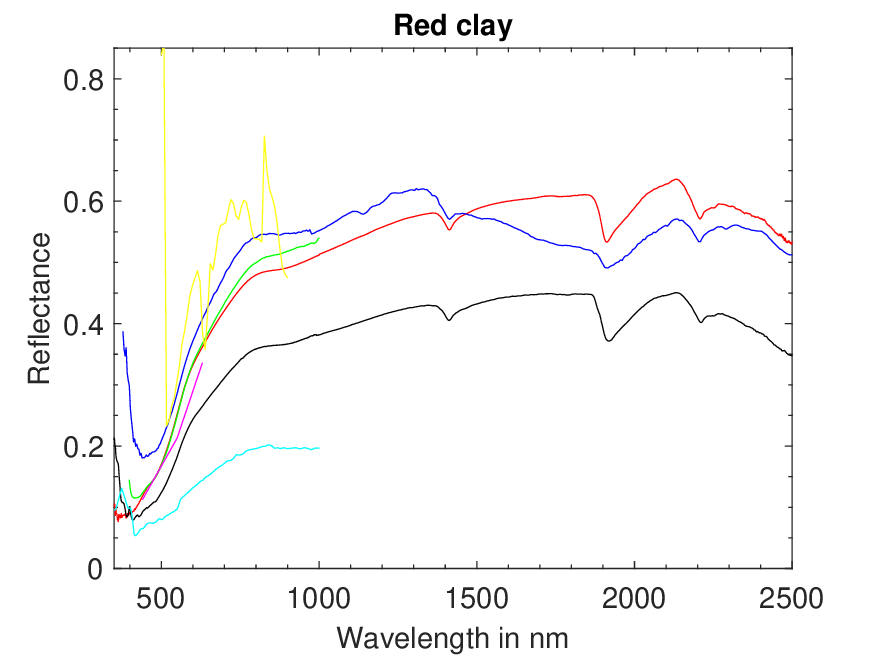}\\ 
 (a) & (b) & (c) \\
 	\includegraphics[width=.32\textwidth]{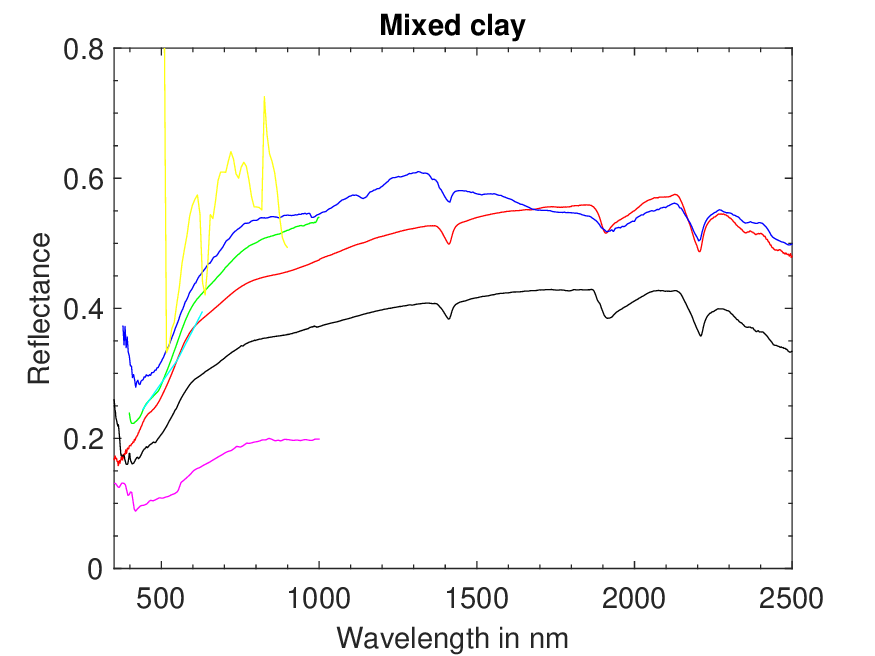} & \includegraphics[width=.32\textwidth]{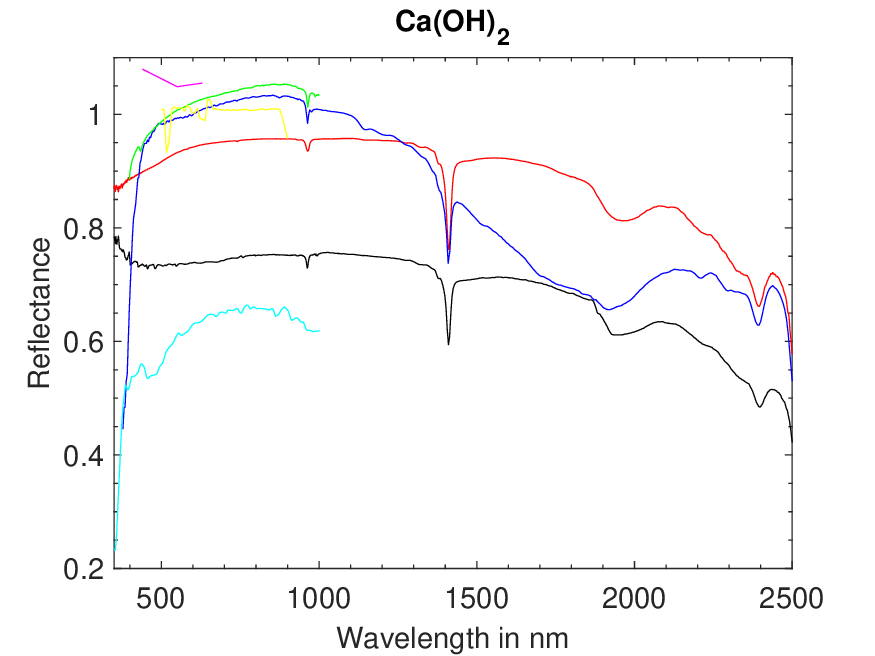} & \includegraphics[width=.32\textwidth]{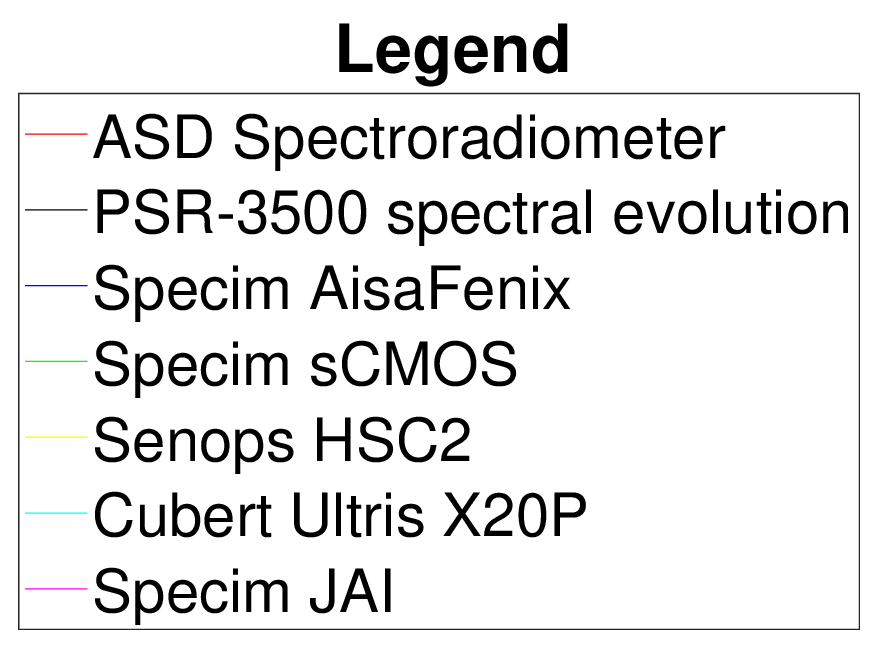}\\
 (d) & (e) \\
  \end{tabular}
 \caption{Spectra of pure clay samples acquired by seven different sensors in the VNIR and SWIR; (a) Kaolin; (b) Roof clay; (c) Red clay; (d) Mixed clay; (e) Ca(OH)$_2$. }
\label{Mineralsspectra}
\end{figure*}

{\color{black} The spectral reflectances of the five pure clay samples acquired by seven different sensors (all but the Cubert panchromatic) are shown in Fig. \ref{Mineralsspectra}.} The clay samples have spectral features around 1400 nm, 1900 nm, and between 2100 and 2500 nm. These features indicate the presence of vibrational hydroxyl processes (\cite{EDucasse2020}). Kaolin (Fig. \ref{Mineralsspectra} (a)) and Ca(OH)$_2$ (Fig. \ref{Mineralsspectra} (e)) are highly reflective. In contrast, roof clay (Fig. \ref{Mineralsspectra} (b)) has the lowest reflectance of the five pure clay samples. Although  the spectra of roof clay (Fig. \ref{Mineralsspectra} (b)), red clay (Fig. \ref{Mineralsspectra} (c)) and mixed clay (Fig. \ref{Mineralsspectra} (d)) show similarities, in general, the overall spectral shape and reflectance values of the pure clay powders are distinctive. This is a prerequisite for accurately estimating the composition of the mixtures.

It is interesting to note that there is a large spectral variability in the acquired spectra. The spectra acquired by the PSR-3500 and the ASD spectroradiometers typically differ by a global scale factor {\color{black} for the same sample.} This variability is likely introduced due to variations in illumination and acquisition angle, and in the distances from the samples to the sensors. In general, band-wise scaling differences can be observed between the acquired spectral reflectances of the different sensors. These effects are caused by variations in illumination and acquisition conditions, the use of different white calibration panels, and specific differences between the sensors.

\subsection{Spectral reflectance of mixtures}

\begin{figure*}[htbp]
\centering
 \begin{tabular}{ccc}
 	\includegraphics[width=.32\textwidth]{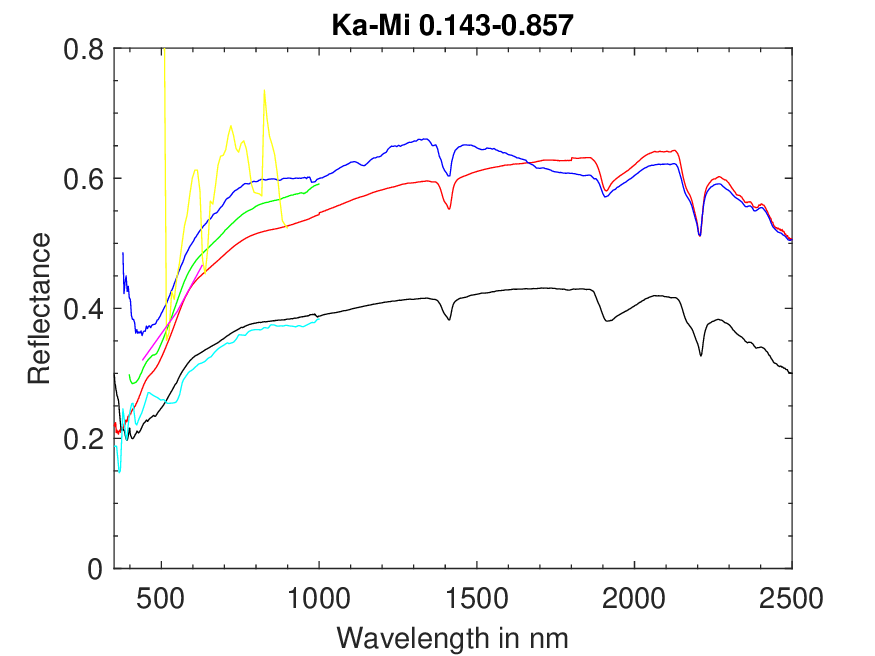} & \includegraphics[width=.32\textwidth]{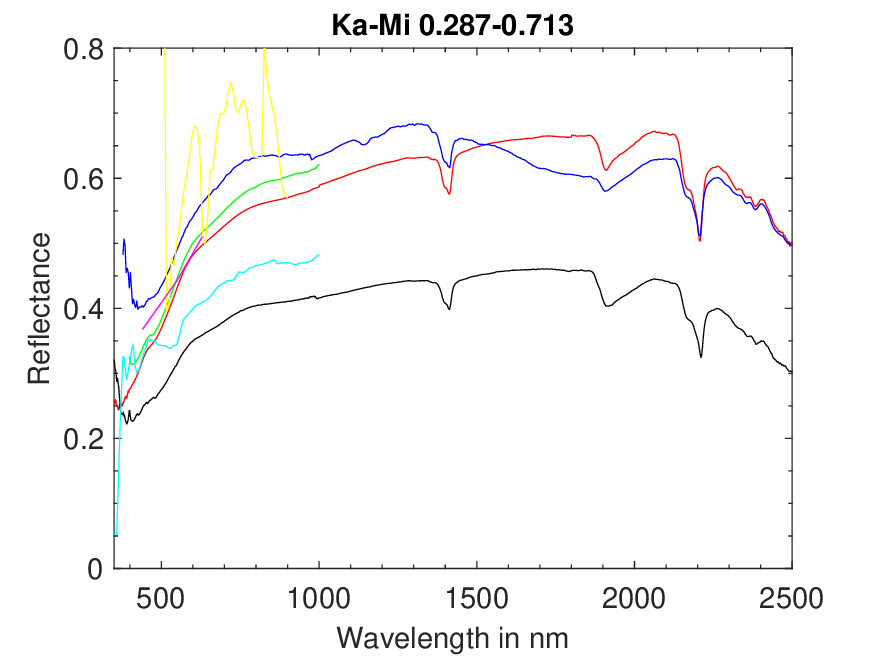} & \includegraphics[width=.32\textwidth]{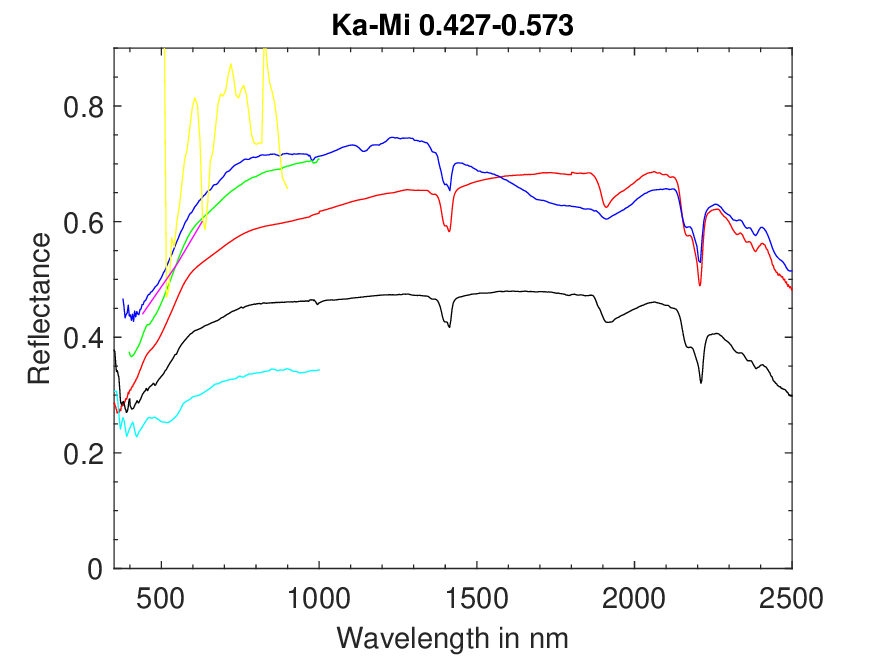}\\ 
 (a) & (b) & (c) \\
	 \includegraphics[width=.32\textwidth]{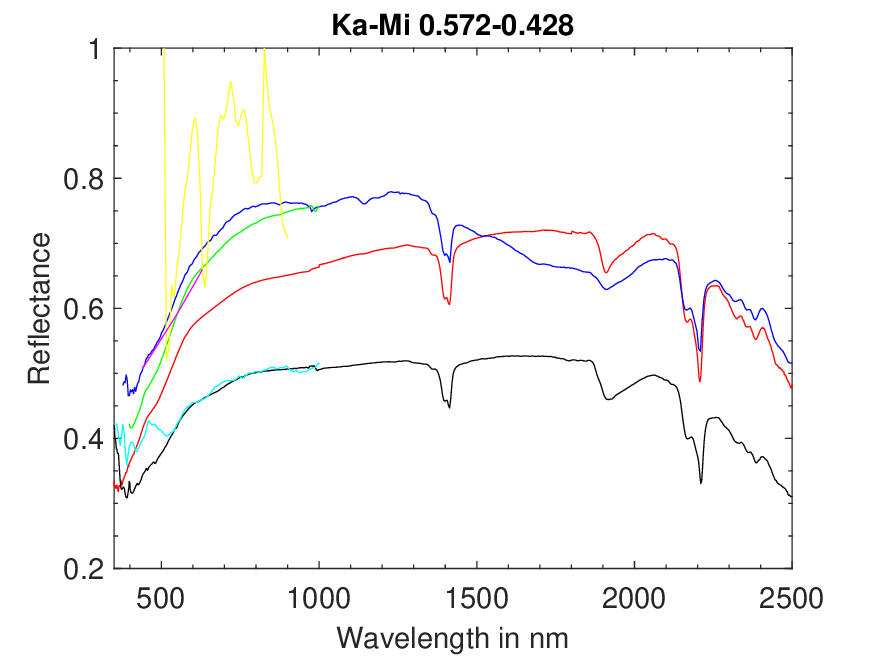}& \includegraphics[width=.32\textwidth]{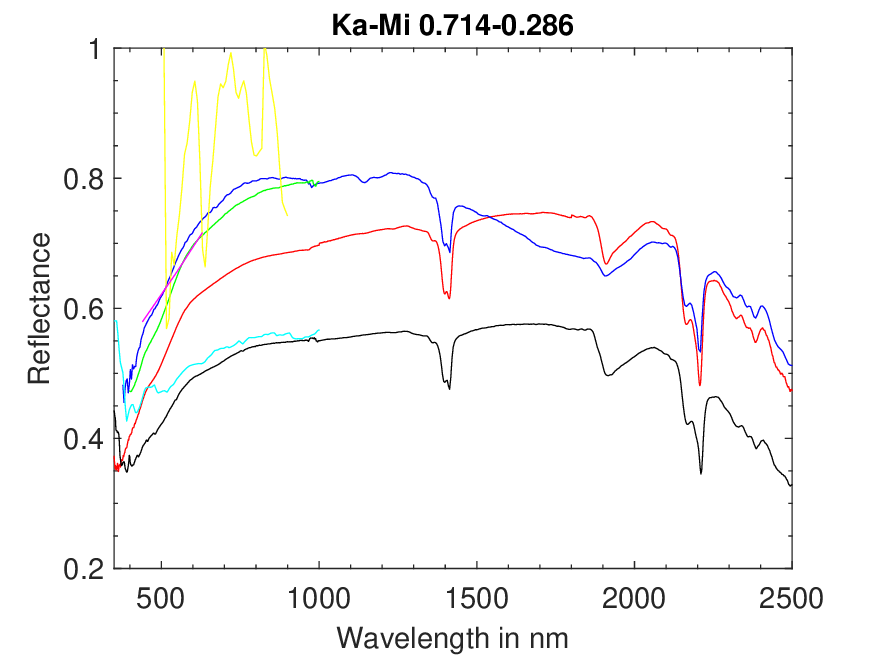}& \includegraphics[width=.32\textwidth]{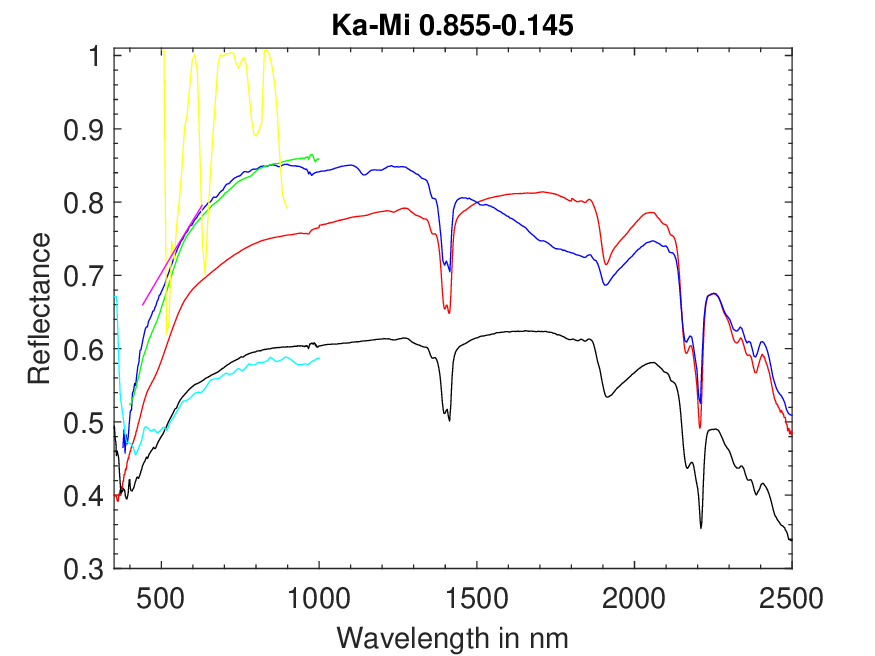}\\
 (d) & (e) & (f) \\
 \includegraphics[width=.22\textwidth]{Pure/Legend_VNIR_SWIR.eps}
  \end{tabular}
 \caption{Spectra of binary mixtures (i.e., a mixture of Kaolin and Mixed clay) acquired by seven different sensors in the VNIR and SWIR; (a) Ka-Mi 14-86; (b) Ka-Mi 28-72; (c) Ka-Mi 43-57; (d) Ka-Mi 57-43; (e) Ka-Mi 72-28; (f) Ka-Mi 86-14. }
\label{Mixturespectra}
\end{figure*}

In Fig. \ref{Mixturespectra}, the spectral reflectance of binary mixtures of Kaolin and Mixed clay, acquired by seven different sensors, is shown. As can be observed, the spectral features of both Kaolin and Mixed clay are present in these spectra, and gradually change when the fractional abundance of each mineral changes in the mixture. For example, the spectral feature of Kaolin around 1400 nm is clearly visible in the sample Ka-Mi 0.855-0.145 and gradually diminishes with a reduction of the abundance of Ka. Again, due to spectral variability, the spectra acquired by different sensors vary widely.

\subsection{Spectral mixture analysis}
It can be assumed that the spectral reflectances obtained from the intimate mixtures are nonlinearly related to the ground truth fractional abundances, due to higher-order scattering of the light rays within the powders before reaching the sensor. To demonstrate the impact of these effects on the abundance estimation, the data are linearly unmixed and the deviations of the linearly estimated fractional abundances from the real ground-truth abundances are studied.

The linear mixing model assumes that the spectral reflectance of a mixture $\mathbf{y}$ is given by:
\begin{equation}
\mathbf{y} = \sum_{j=1}^p a_j \mathbf{e}{_j} + \mathbf{n},
\end{equation}
where $\mathbf{e}{_j}$ are the endmember spectra of the pure clays, $a_j$ is the fractional abundance of endmember $j$ and  $\mathbf{n}$ is gaussian noise. To be physically interpretable, the fractional abundances are generally assumed to be non-negative and sum-to-one. The fully constrained least squares unmixing method (FCLSU) minimizes $\norm{\mathbf{y}-\sum_{j=1}^p {a}_{j}\mathbf{e_j}}^2$ s.t. $\sum_{j} a_{j}=1$, $\forall j: a_{j} \geq 0$.

\begin{figure*}[htbp]
\centering
 \begin{tabular}{ccc}
 	\includegraphics[width=.32\textwidth]{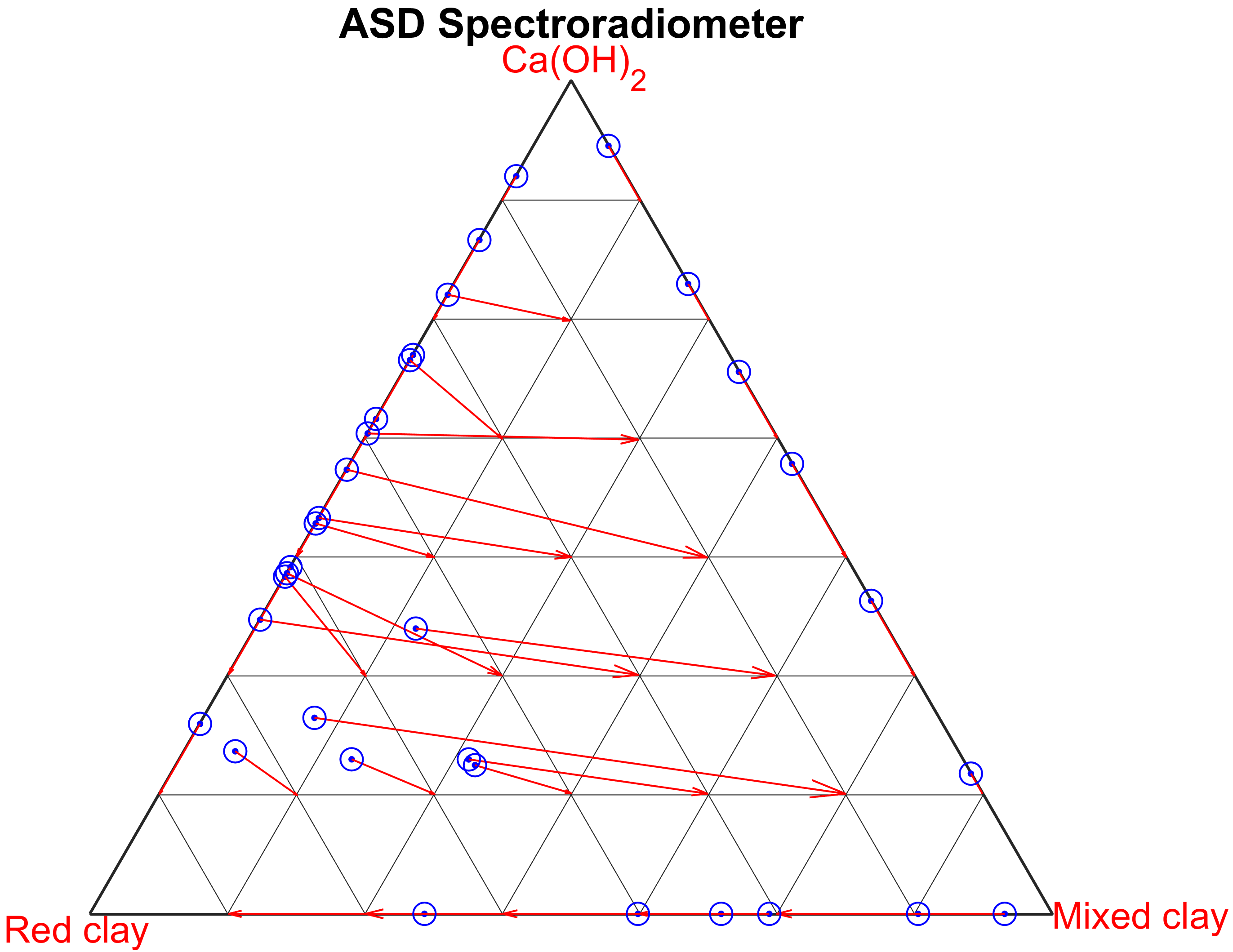}& \includegraphics[width=.32\textwidth]{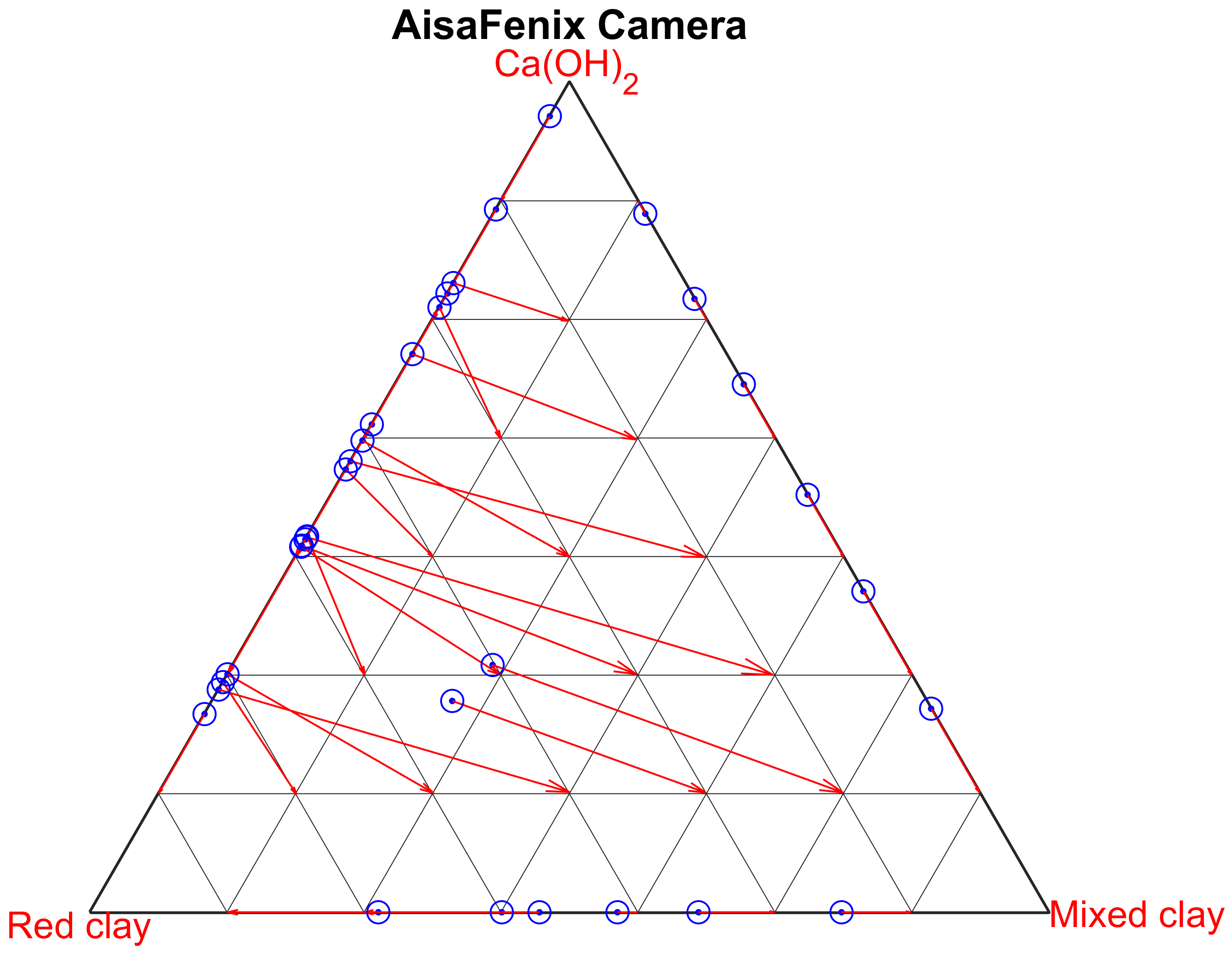} & \includegraphics[width=.32\textwidth]{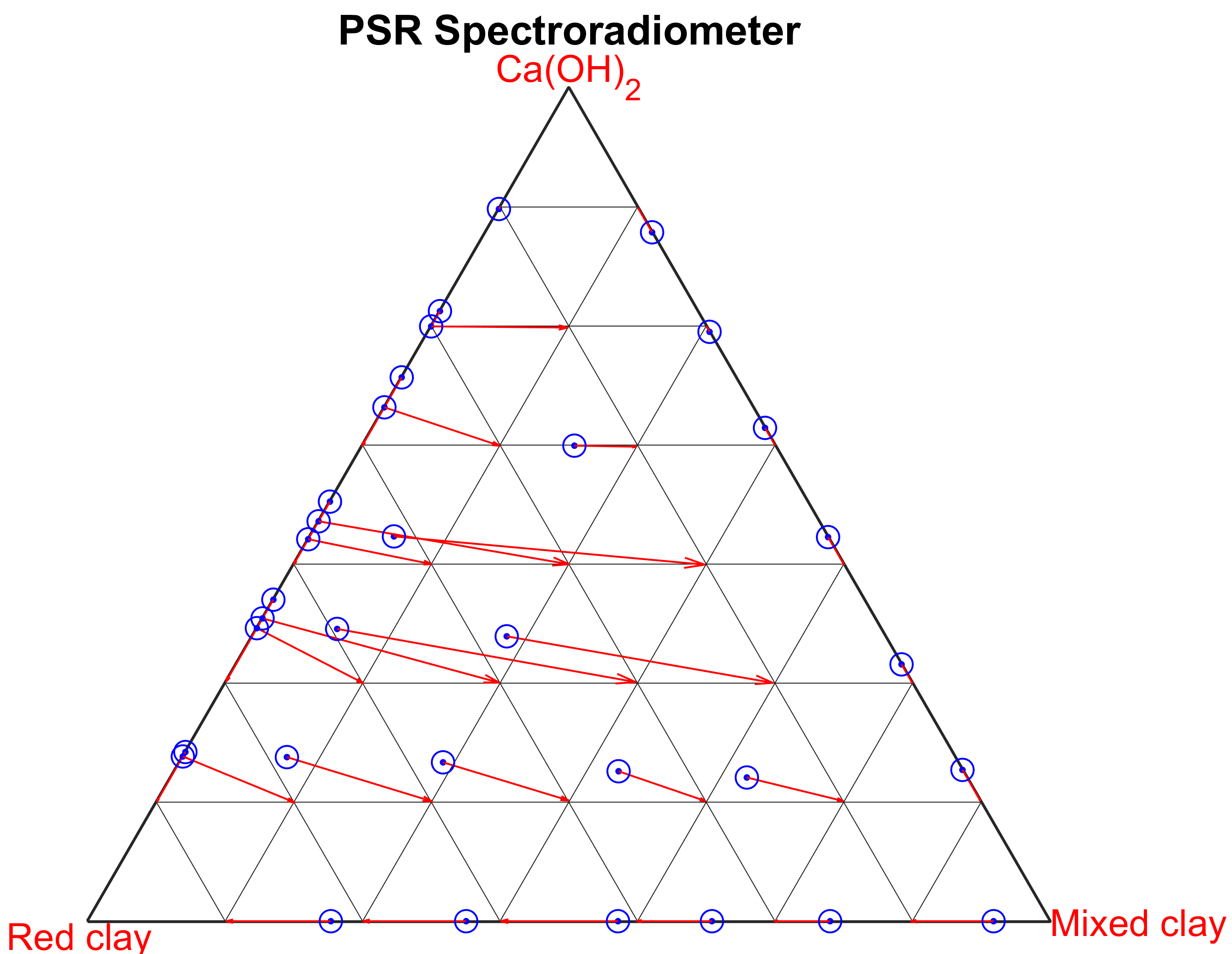}\\ 
 (a) & (b) & (c) \\
  \includegraphics[width=.32\textwidth]{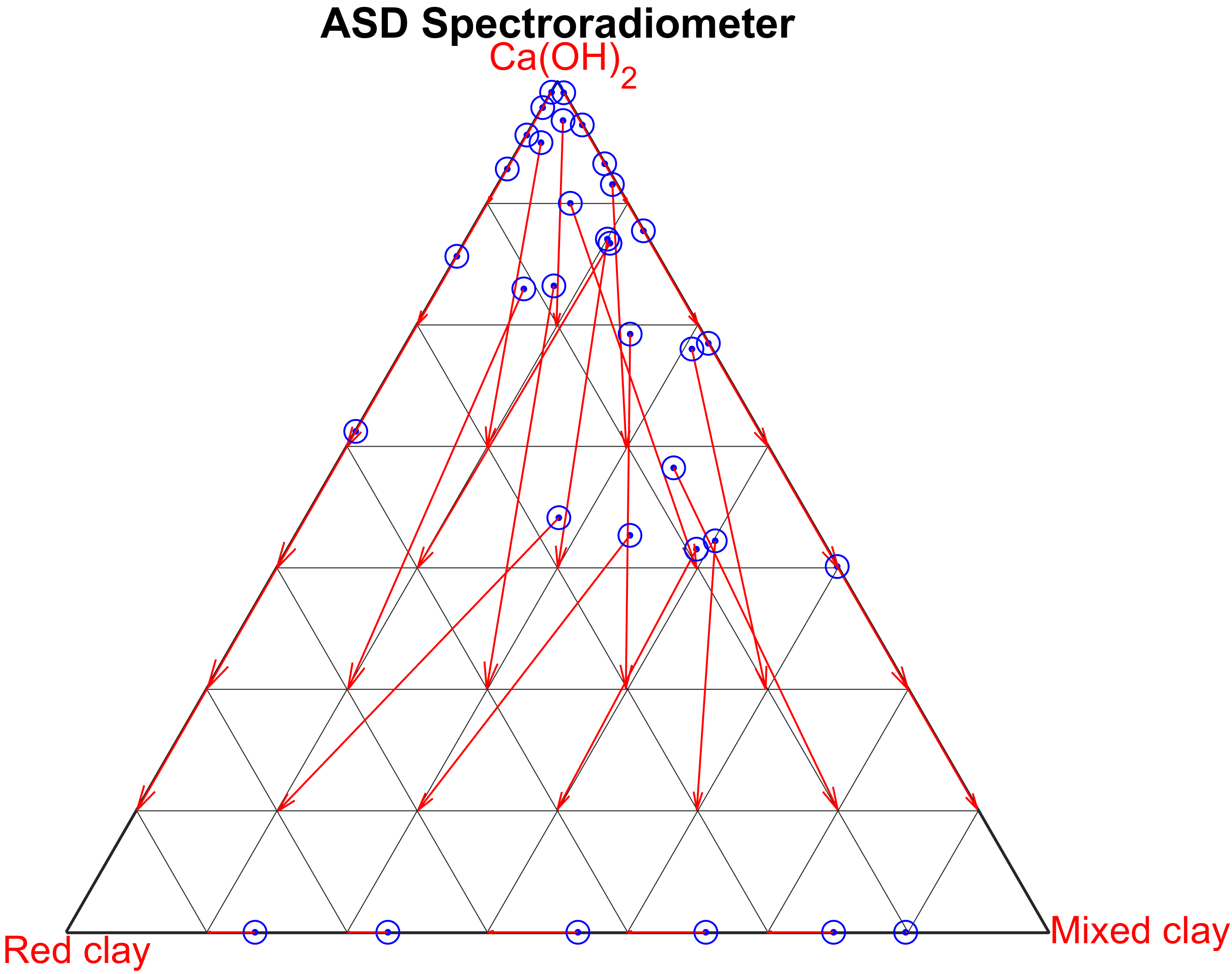}& \includegraphics[width=.32\textwidth]{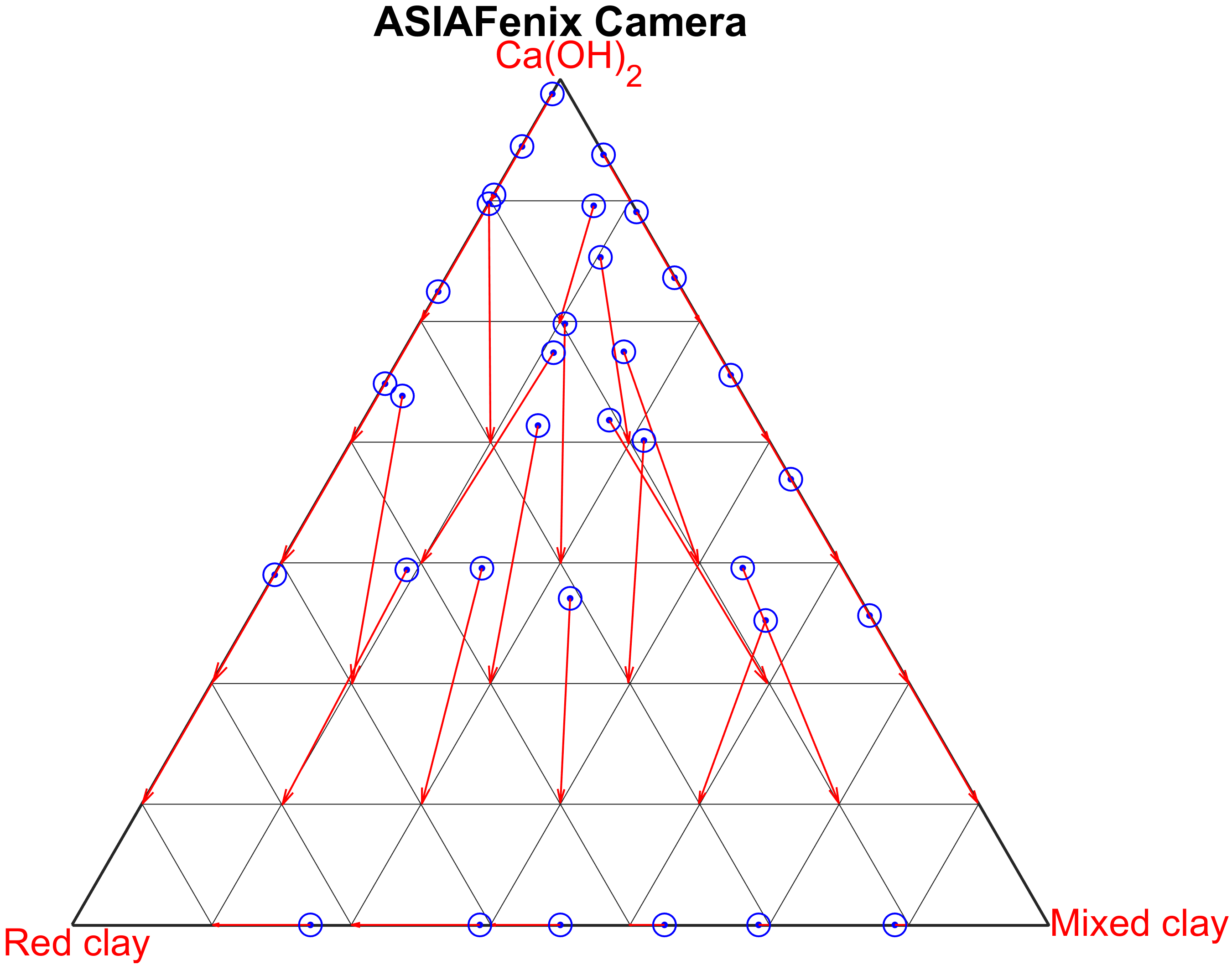} & \includegraphics[width=.32\textwidth]{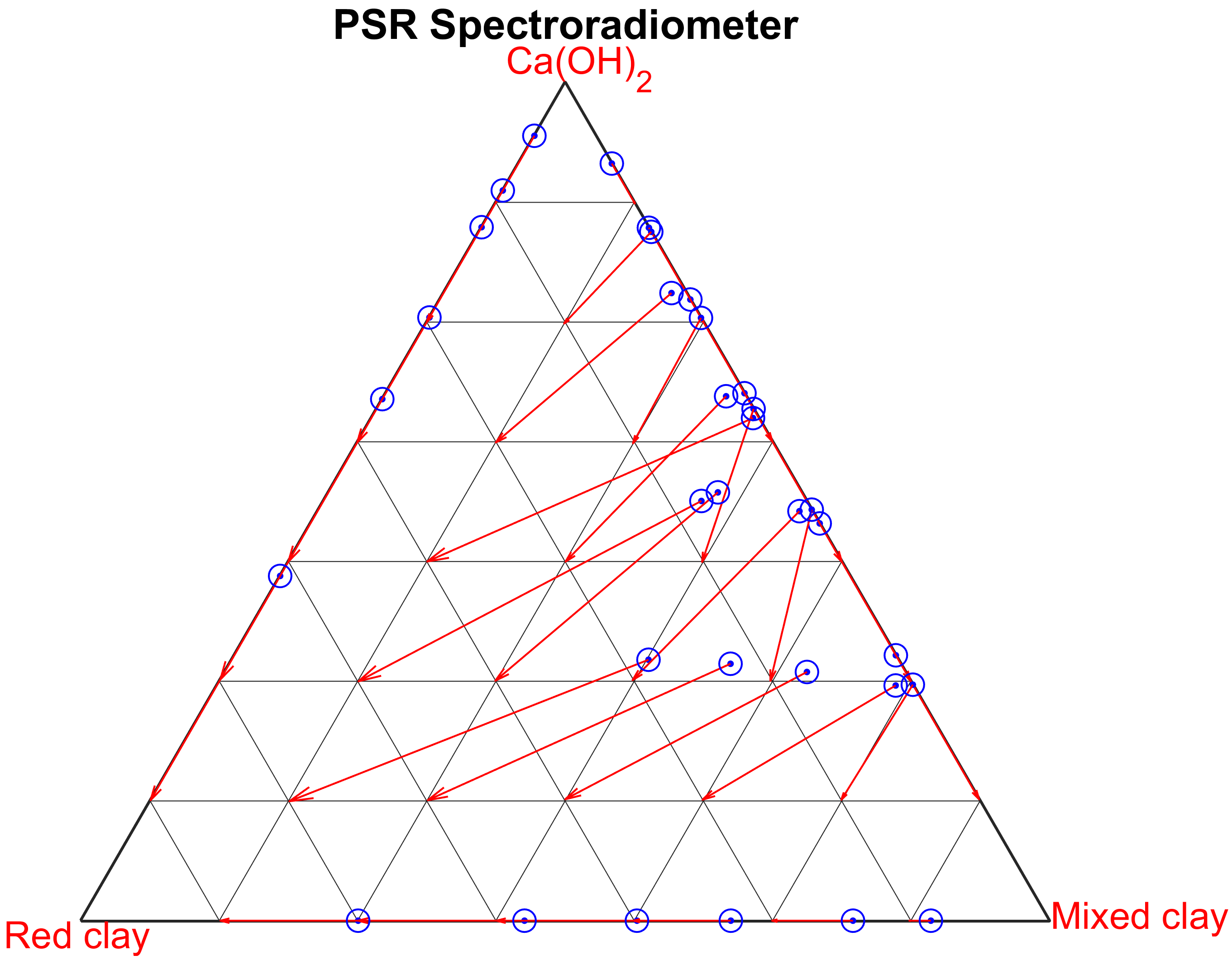}\\
  (d) & (e) & (f) \\
  \end{tabular}
 \caption{{\color{black} Unmixing results overlaid on the ternary diagram of the mixtures of three clays; Top row: Linear unmixing; Bottom row: Hapke unmixing. In the figure, the blue dots denote the estimated positions of the mixtures while the red arrows show the real positions of the mixtures; (a and d) ASD Spectroradiometer; (b and e) AisaFenix Camera; (c and f) PSR Spectroradiometer.}}
\label{Unmixing_results}
\end{figure*}

In Fig. \ref{Unmixing_results}, Top row, we display the estimated fractional abundances by FCLSU on the reflectance dataset in the VNIR/SWIR, obtained from the binary and ternary mixtures of the Red Clay, Mixed Clay, and Calcium hydroxide, overlaid on the ternary diagram of the clay mixtures. 

{\color{black} For comparison, the estimated fractional abundances by the Hapke model (Fig. \ref{Unmixing_results}, Bottom row) are shown as well. The Hapke model estimates the fractional abundances from the spectral reflectance by minimizing the following equation: $\norm{\mathbf{y}-\frac{\mathbf{W a}}{\left(1+2\textup{cos}(\theta_e)\sqrt{1-\mathbf{W a}} \right)\left(1+2\textup{cos}(\theta_i)\sqrt{1-\mathbf{W a}} \right)}}^2$, s.t. $\sum_{j} a_{j}=1$, $\forall j: a_{j} \geq 0$, where $\theta_i$ and $\theta_e$ are the angles with the normal of the incoming and outgoing radiation respectively and $\mathbf{W}$ are the single scattering albedos of the endmembers. 

In the figure, the blue dots denote the estimated abundances, while the red arrows show the real position of the mixtures in the ternary diagram. As can be observed, the error in the estimated fractional abundances of the binary mixtures is significant. Moreover, both the linear and the Hapke model project many of the ternary mixtures onto the faces of the simplex leading to a significant error in the estimated fractional abundances. }

\section{Data acquisition in MWIR and LWIR wavelength regions}
\label{MWIR and LWIR sensors}
\subsection{Handheld sensor}
\begin{figure}[htbp]
\centering
 \begin{tabular}{l}
 	\includegraphics[width=.3\textwidth]{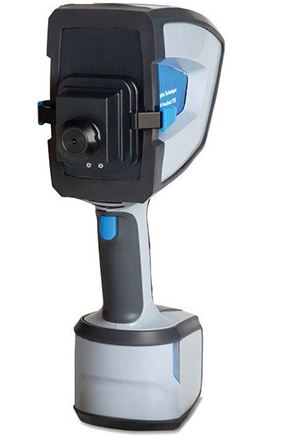} 
  \end{tabular}
 \caption{Agilent 4300 FTIR.}
\label{Agilent}
\end{figure}

Spectral reflectances of all 330 samples were acquired in the region (2500-15385) nm by the Agilent 4300 Handheld Fourier transform infrared (FTIR) Spectrometer (Fig. \ref{Agilent}). An FTIR has an excitation source that illuminates the sample. The sensor contains an interferometer (a configuration of 2 mirrors) that periodically blocks or passes each wavelength of the incoming light by moving one of the two mirrors. The raw data collected by this sensor is often referred to as an interferogram. This raw data is then converted onto reflectance by applying Fourier transform. 

\subsection{Imaging sensors}
{\color{black} Hyperspectral images are acquired from all samples in the MWIR and LWIR wavelength regions by four different cameras: Specim FX50, Telops MWIR, Specim AisaOwl, and Telops HyperCam (see Table \ref{Sensor summary} for detailed information)}. Both the Specim FX50 and the Specim AisaOwl were mounted on the drill-core scanner (Fig. \ref{Drill core scanner}) while the Telops MWIR was mounted on the conveyor belt (Fig.\ref{Conveyor belt}). The distance between the Specim FX50 and the moving table is approximately 0.85 m, while the Specim AisaOwl is approximately 1.06 m away from the moving table. The Telops HyperCam  has its own scanning platform. Due to this camera's limited field of view (14$^0$ x 11$^0$), each sample was scanned individually. As with the VNIR and SWIR cameras, a limited number of pixels was selected from the center of the sample using a rectangular bounding box. The final image size of each sample is 8 $\times$ 8 pixels for the Specim FX50, 8 $\times$ 8 pixels for the Specim AisaOwl, and 18 $\times$ 18  pixels for the Telops HyperCam. All spectra are averaged over the entire sample image. Due to the sparse noise in data acquired by Specim AisaOwl, we applied a mixed noise removal technique called hyperspectral mixed Gaussian and sparse noise reduction on the dataset \cite{BRasti2020}.


\subsection{Spectral reflectance of pure clay samples}
\begin{figure*}[htbp]
\centering
 \begin{tabular}{ccc}
 	\includegraphics[width=.32\textwidth]{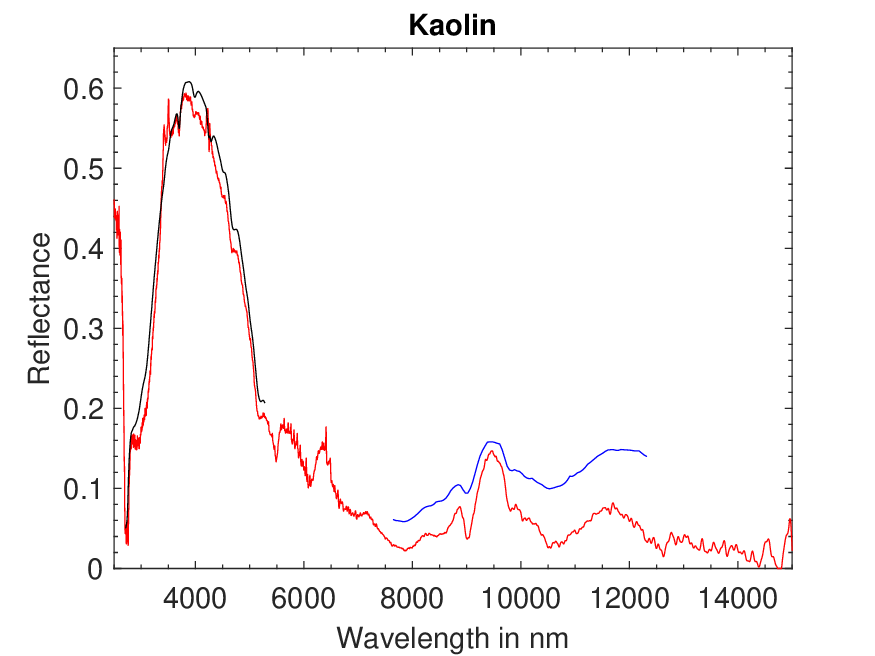}& \includegraphics[width=.32\textwidth]{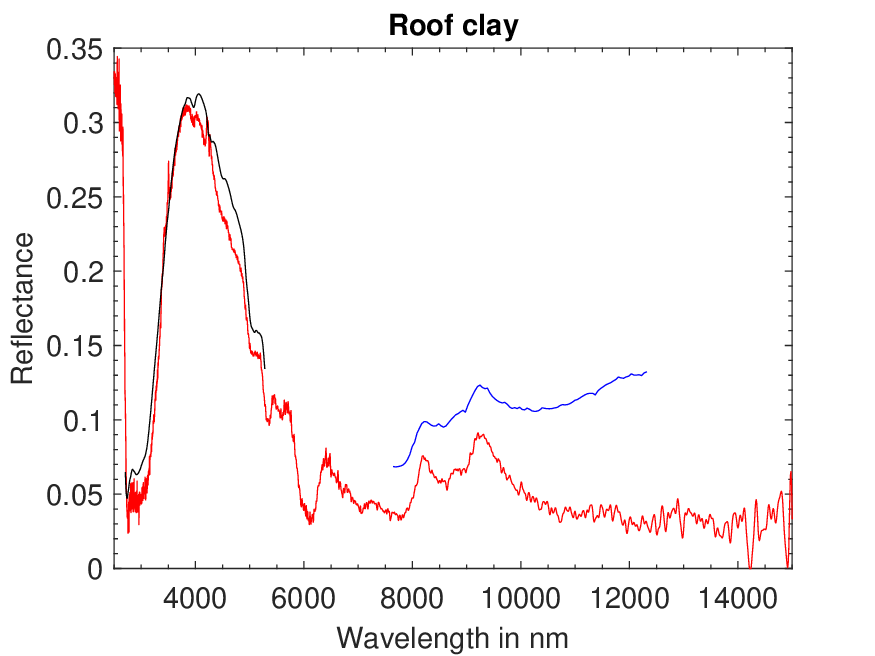}& \includegraphics[width=.32\textwidth]{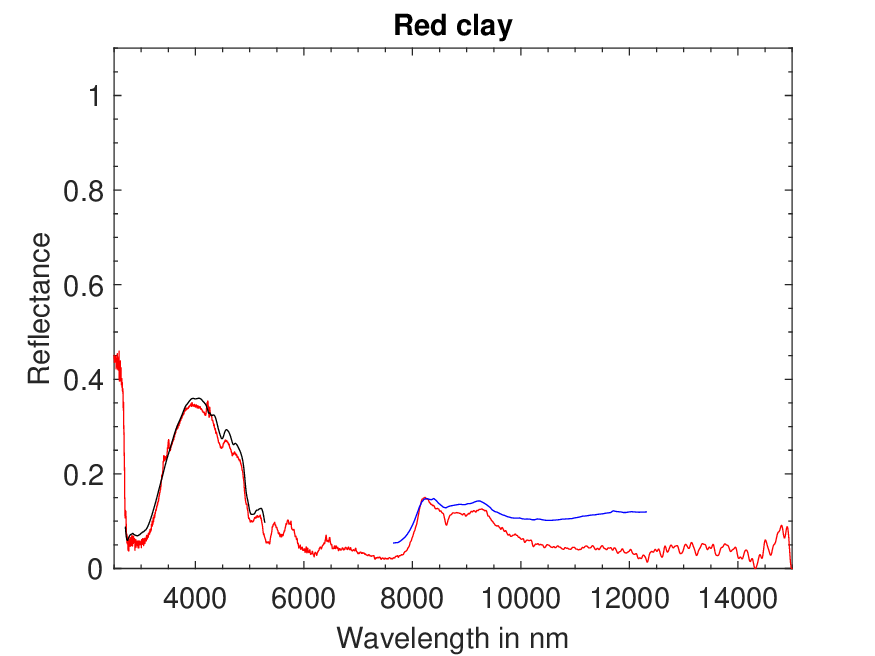}\\ 
 (a) & (b) & (c) \\
 \includegraphics[width=.32\textwidth]{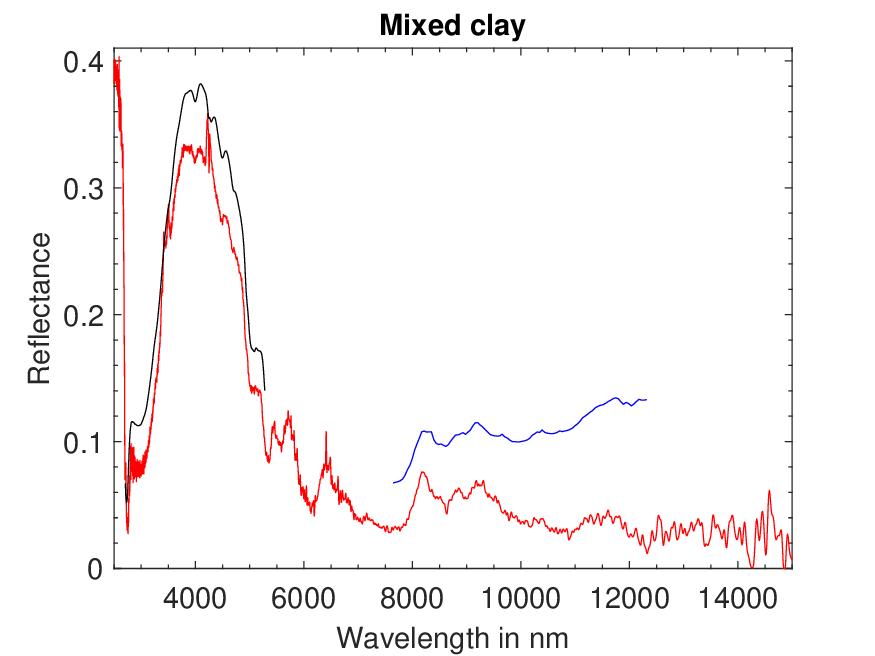}& \includegraphics[width=.32\textwidth]{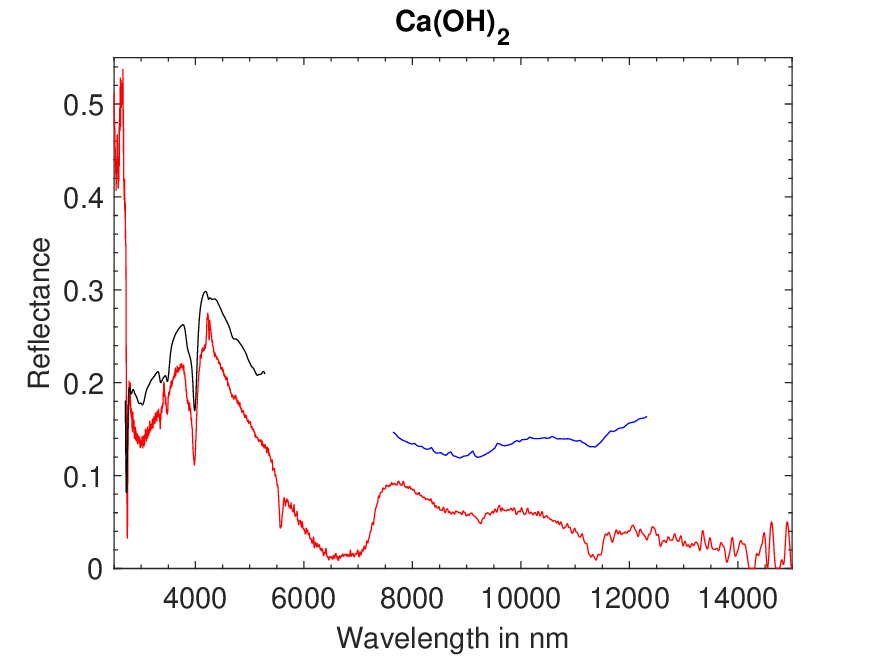} & \includegraphics[width=.22\textwidth]{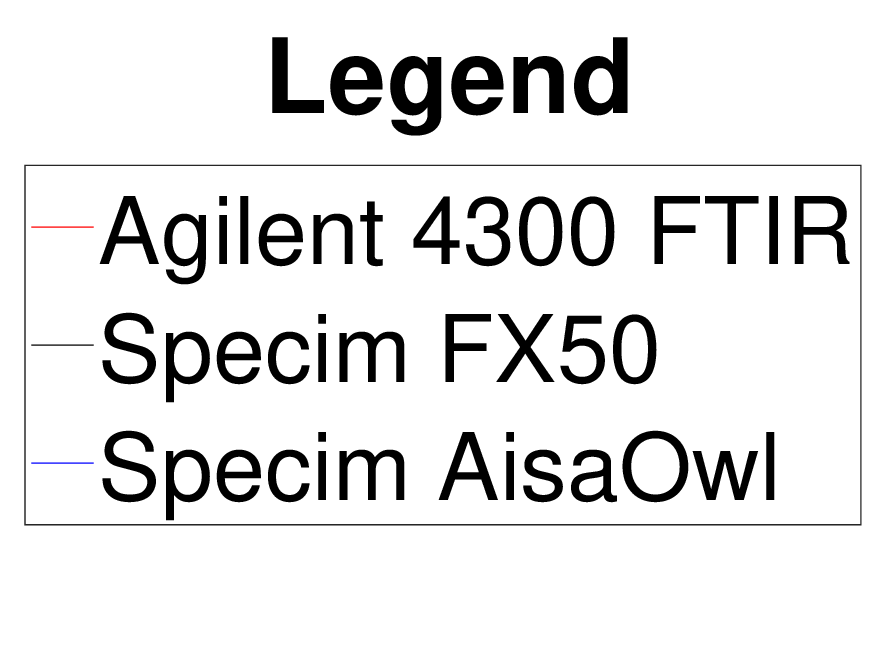}\\
 (d) & (e) \\
  \end{tabular}
 \caption{Spectra of pure clay samples acquired by three different sensors in the MWIR and LWIR;  (a) Kaolin; (b) Roof clay; (c) Red clay; (d) Mixed clay; (e) Ca(OH)$_2$. }
\label{MineralsspectraMWIRLWIR}
\end{figure*}

Fig. \ref{MineralsspectraMWIRLWIR} shows the  spectral reflectance of the pure clay samples acquired by three different sensors: Agilent 4300 FTIR, Specim FX50, and Specim AisaOwl (datasets acquired by Telops MWIR and Telops HyperCam are not shown here due to low-quality spectral information). Most clay minerals have multiple absorption bands in the MWIR and LWIR (2500 nm to 25000 nm) that can be related to the fundamental stretching and bending vibrations of their fundamental functional groups, e.g., the OH and Si-O groups \cite{MADEJOVA2017107}. Apart from an intensity difference, Kaolin, Roof clay, Red clay, and Mixed clay have the same spectral shape between 2500 and 6000 nm, {\color{black} while the Ca(OH)$_2$ sample shows a significantly different spectral shape (around 4000 nm) due to the presence of a large amount of calcium carbonate. In general, the noise levels for the LWIR appear to be greater than in the VNIR/SWIR domains, especially after 10000 nm.} 
Unlike the VNIR/SWIR, there is not much spectral variability in the MWIR/LWIR. The spectra of the pure clay powders acquired by the Agilent 4300 Handheld FTIR and the Specim FX50 are almost identical, while the spectra acquired by the Specim AisaOwl differ only slightly.

\subsection{Spectral reflectance of mixtures}
\begin{figure*}[htbp]
\centering
 \begin{tabular}{ccc}
 	\includegraphics[width=.32\textwidth]{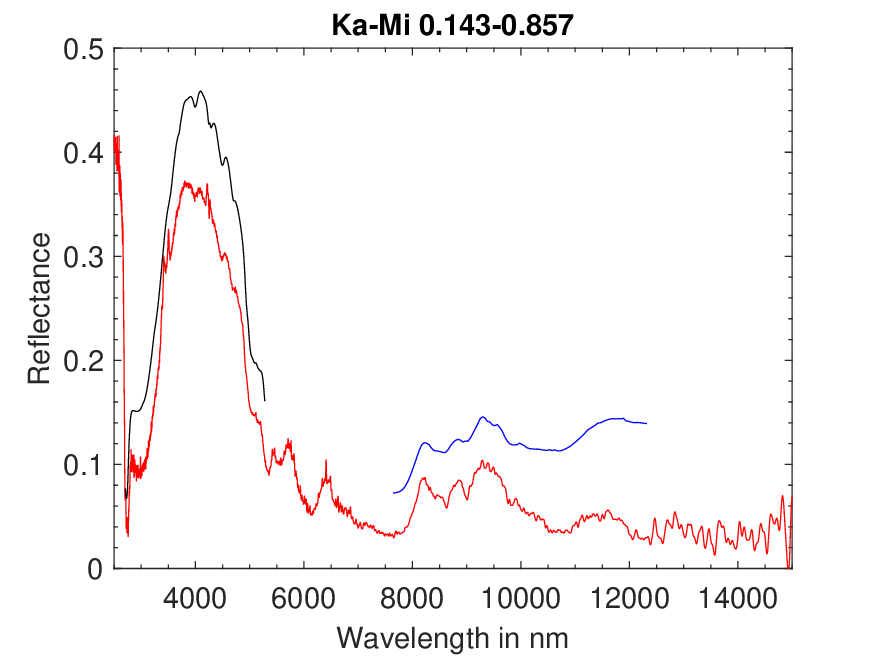}& \includegraphics[width=.32\textwidth]{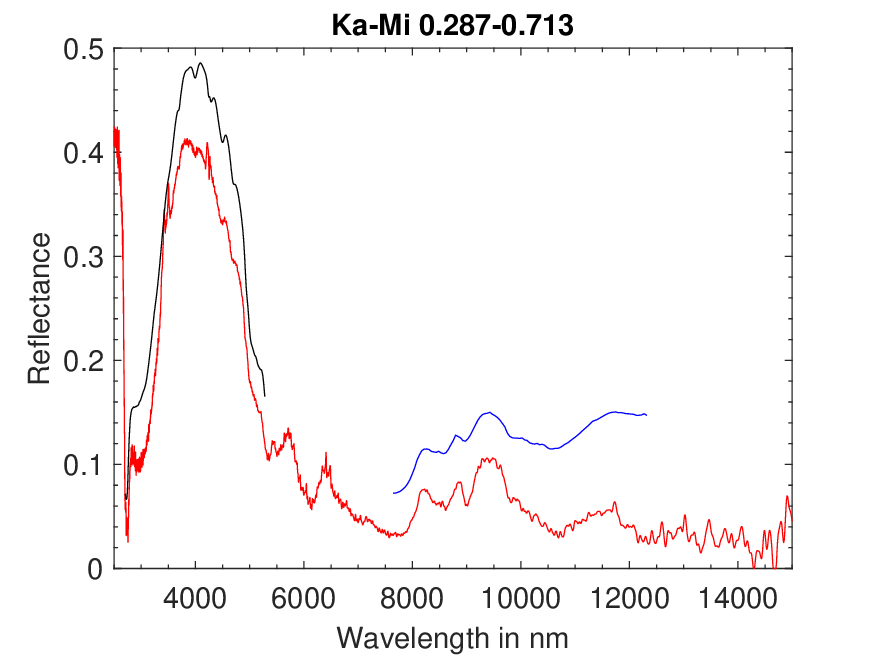}& \includegraphics[width=.32\textwidth]{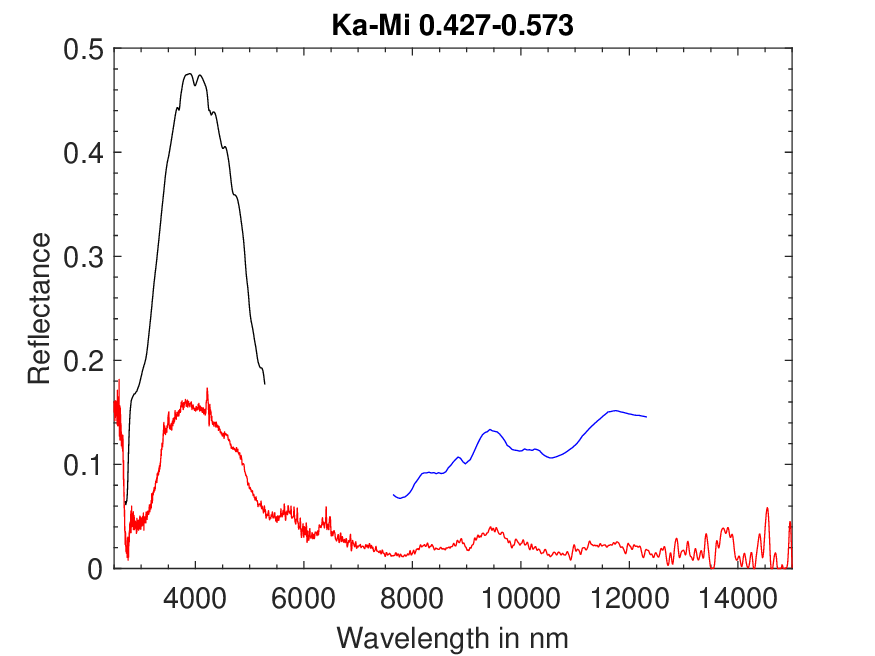}\\ 
 (a) & (b) & (c) \\
	 \includegraphics[width=.32\textwidth]{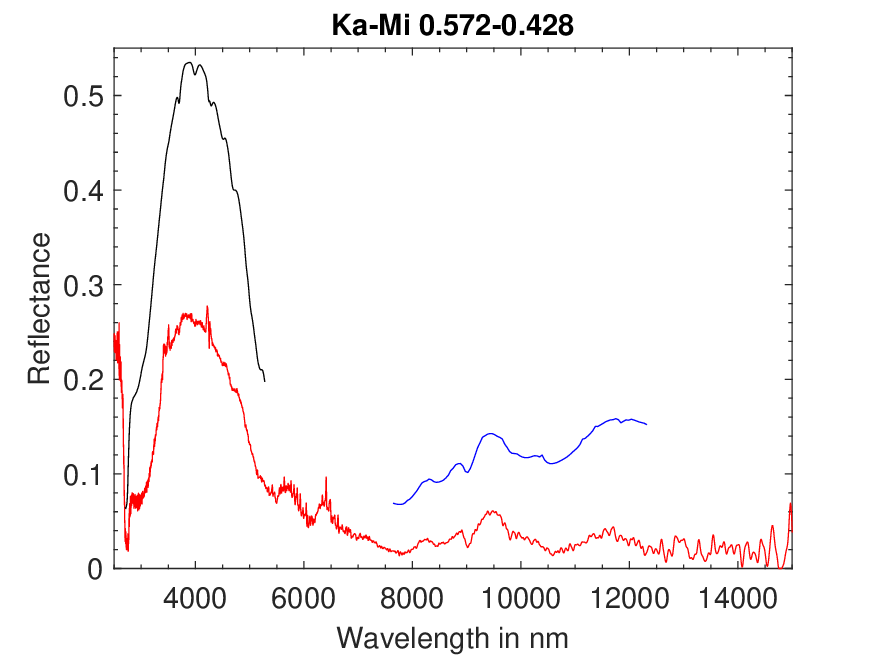}& \includegraphics[width=.32\textwidth]{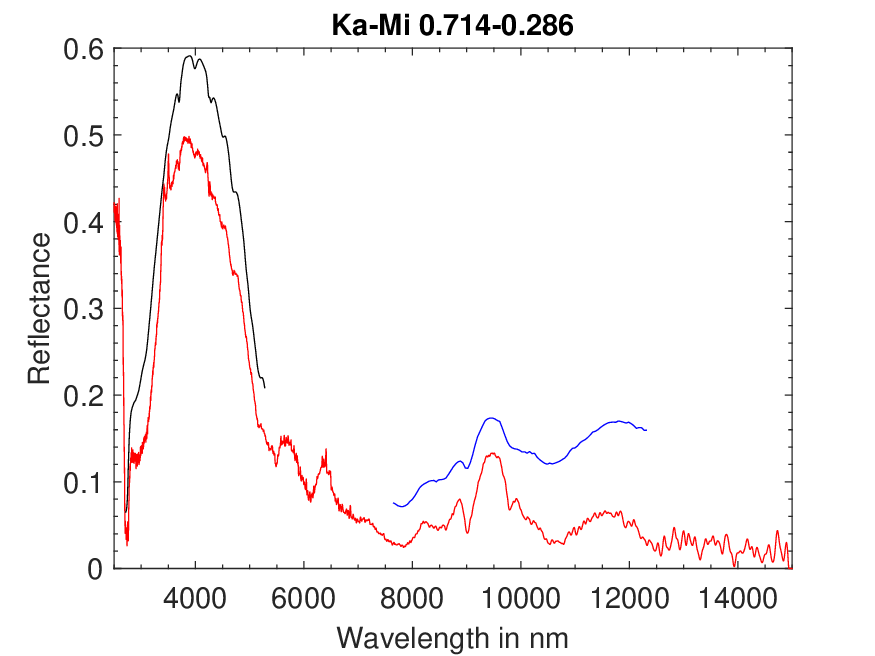}& \includegraphics[width=.32\textwidth]{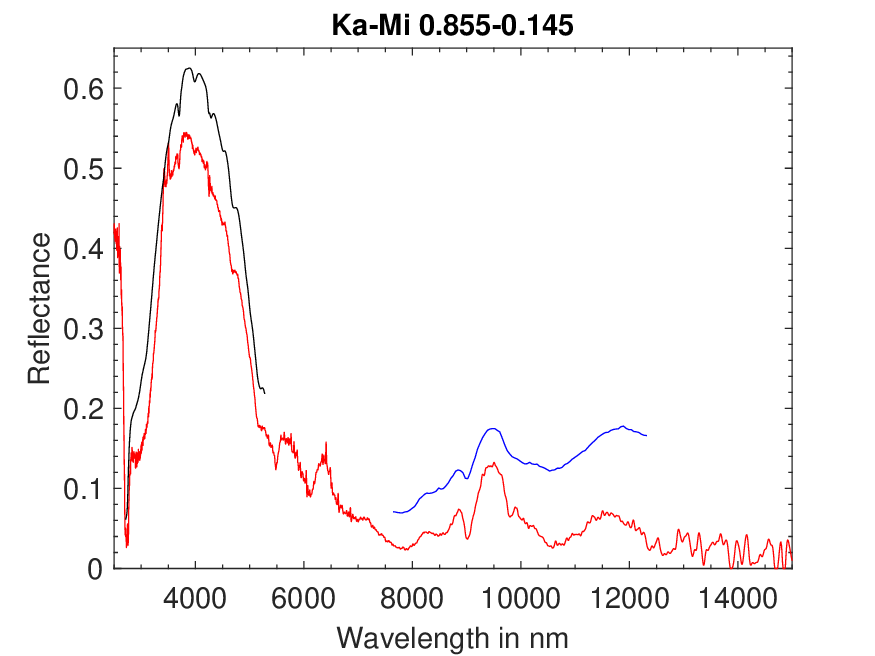}\\
 (d) & (e) & (f) \\
 \includegraphics[width=.22\textwidth]{Pure/Legend_MWIR.eps}
  \end{tabular}
 \caption{Spectra of binary mixtures (i.e., a mixture of Kaolin and Mixed clay) acquired by three different sensors in the MWIR and LWIR; (a) Ka-Mi 14-86; (b) Ka-Mi 28-72; (c) Ka-Mi 43-57; (d) Ka-Mi 57-43; (e) Ka-Mi 72-28; (f) Ka-Mi 86-14. }
\label{MixturespectraMWIR}
\end{figure*}

Fig. \ref{MixturespectraMWIR} shows the spectral reflectance of the binary mixtures of Kaolin and Roof clay, acquired by three different sensors in the MWIR and LWIR. The spectral variability is much larger for the binary mixtures than for the pure materials.

\subsection{Spectral mixture analysis}
\begin{figure*}[htbp]
\centering
 \begin{tabular}{ccc}
 	\includegraphics[width=.32\textwidth]{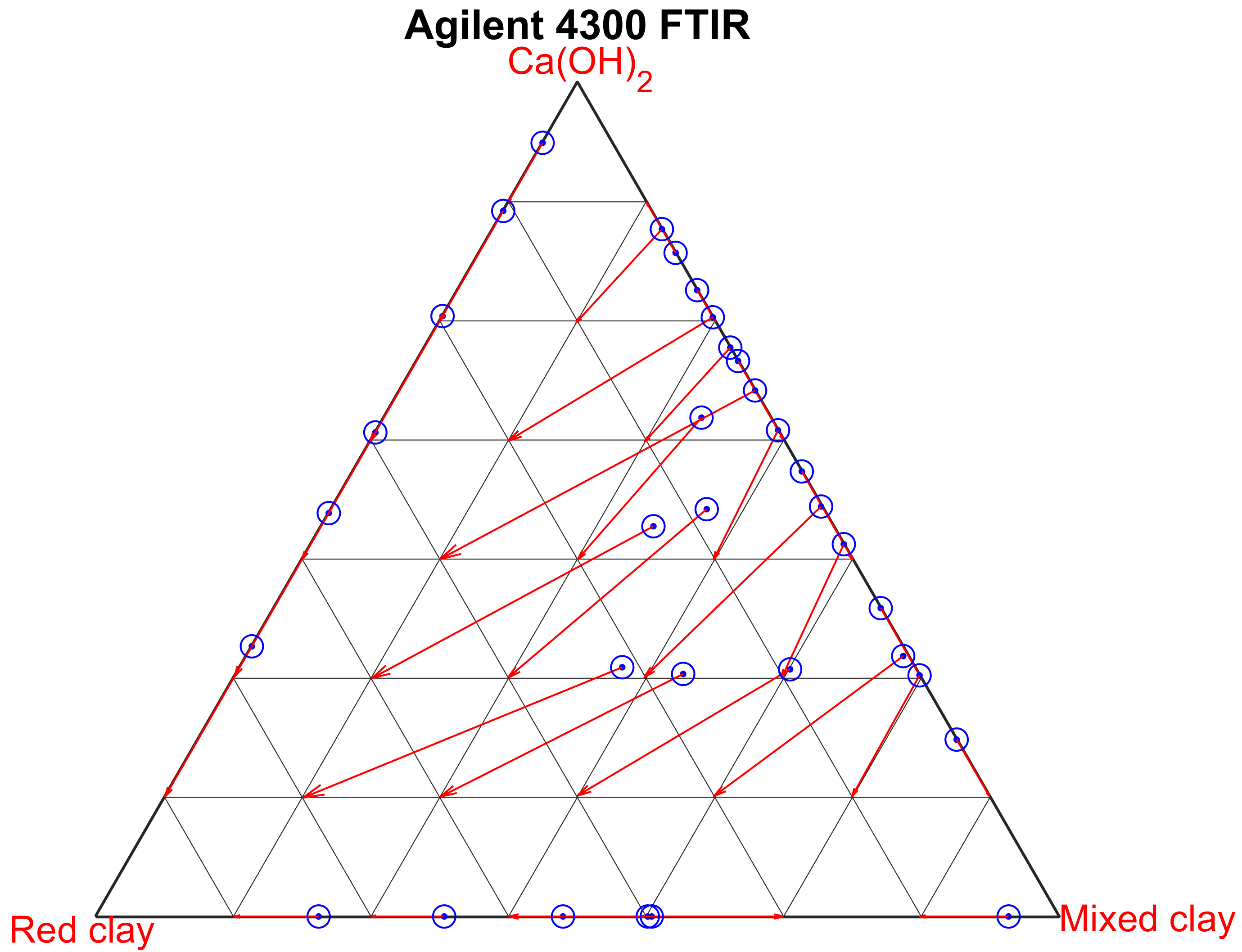}& \includegraphics[width=.32\textwidth]{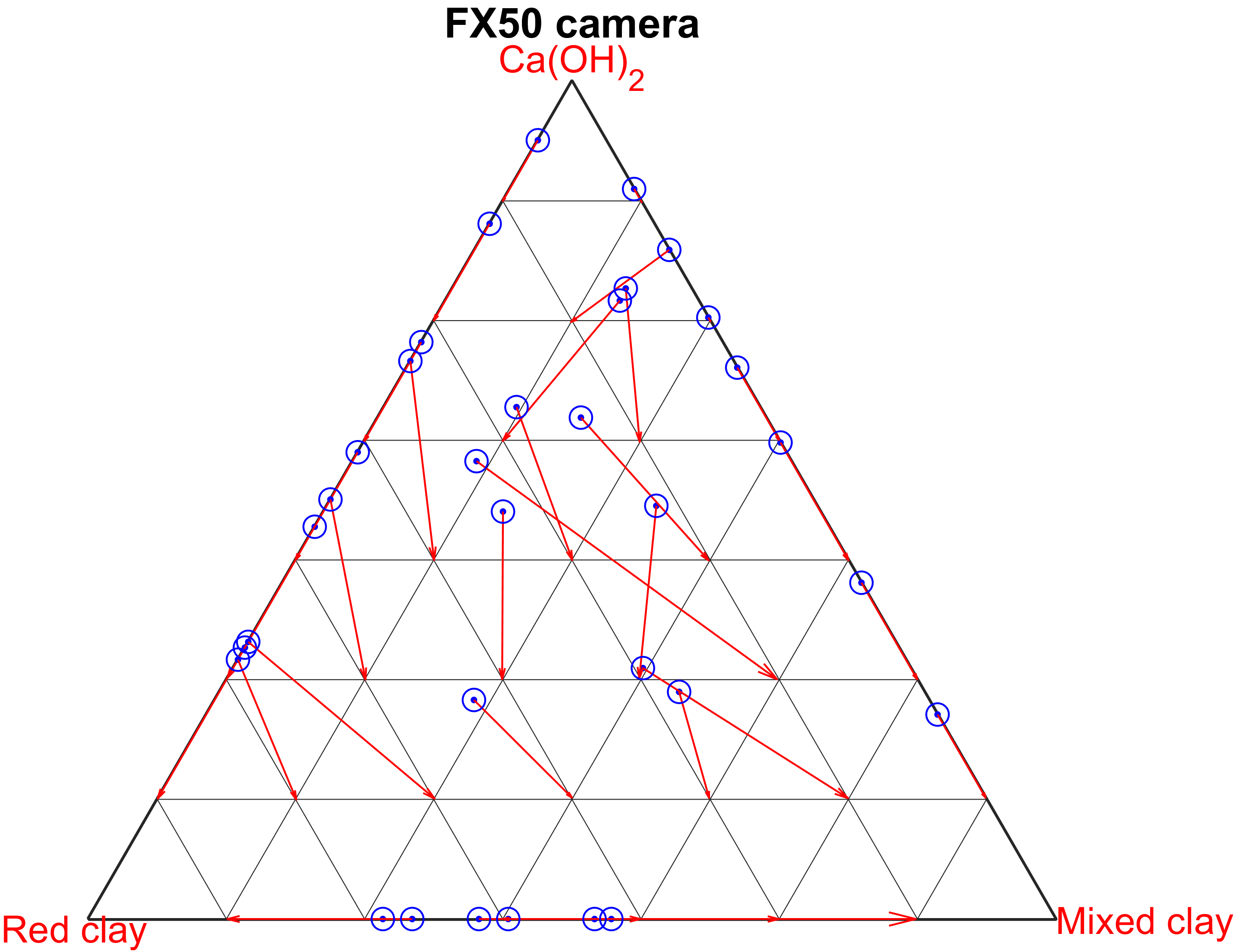}& \includegraphics[width=.32\textwidth]{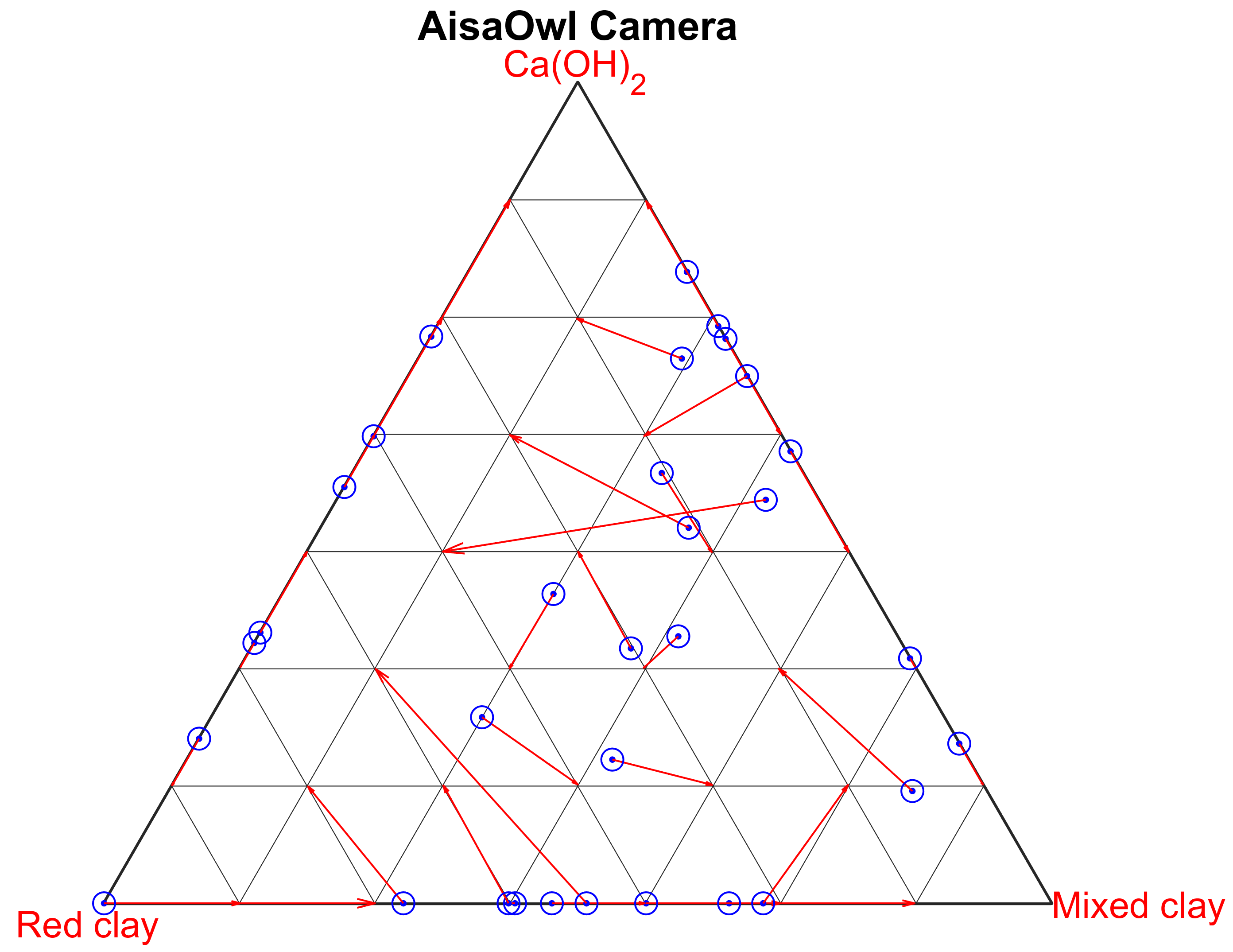}   \\
 (a) & (b) & (c) \\
 	 \includegraphics[width=.32\textwidth]{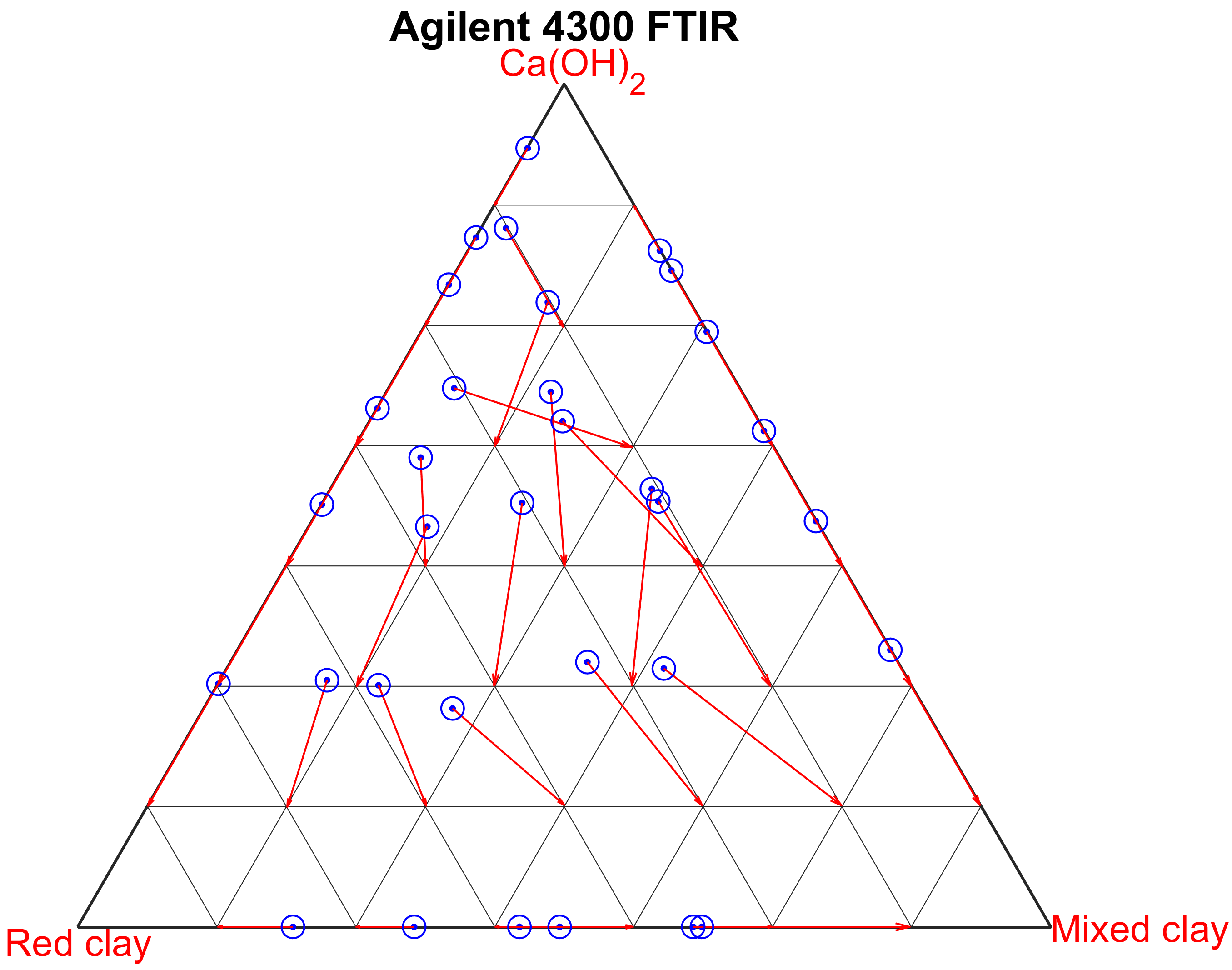}& \includegraphics[width=.32\textwidth]{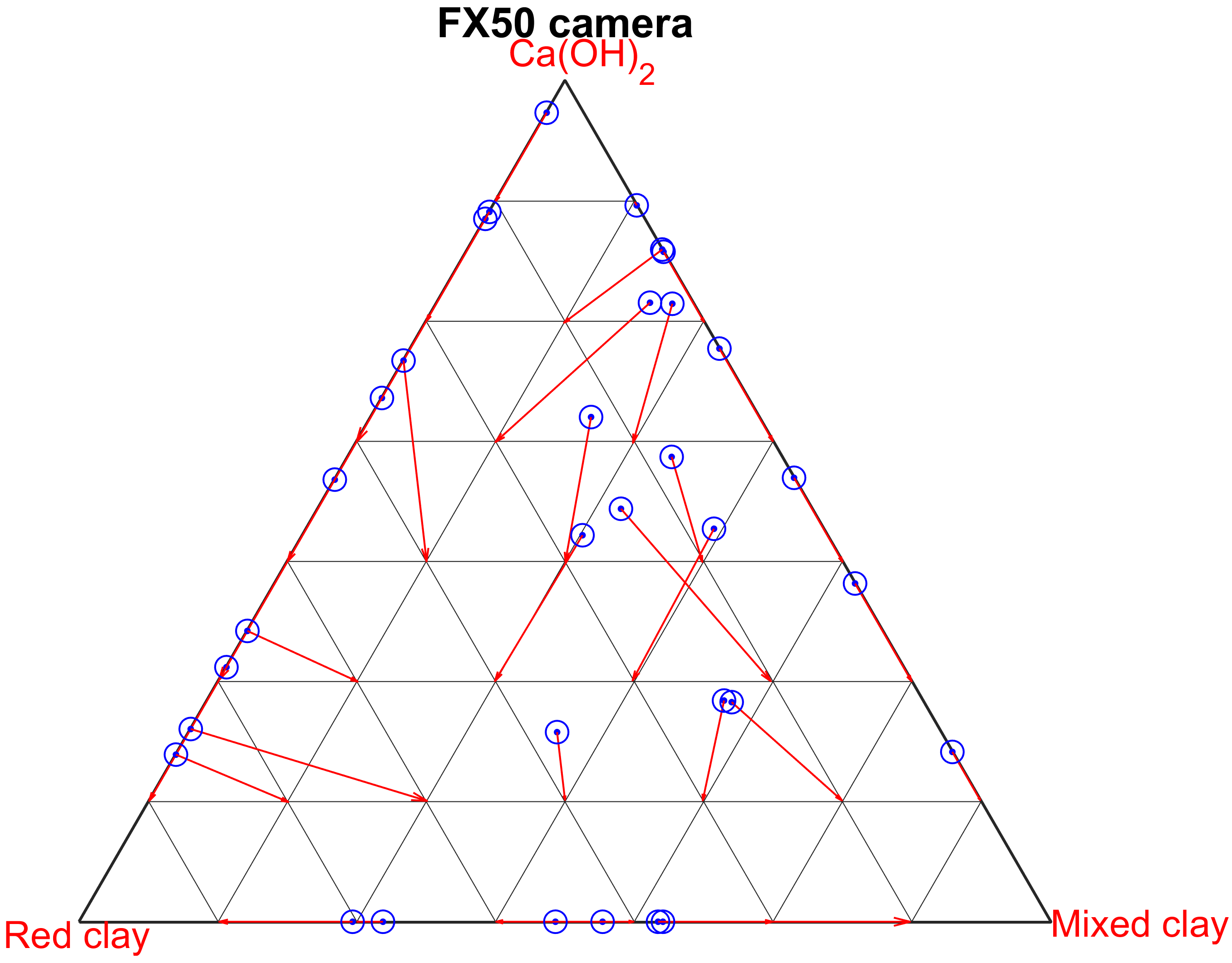}& \includegraphics[width=.32\textwidth]{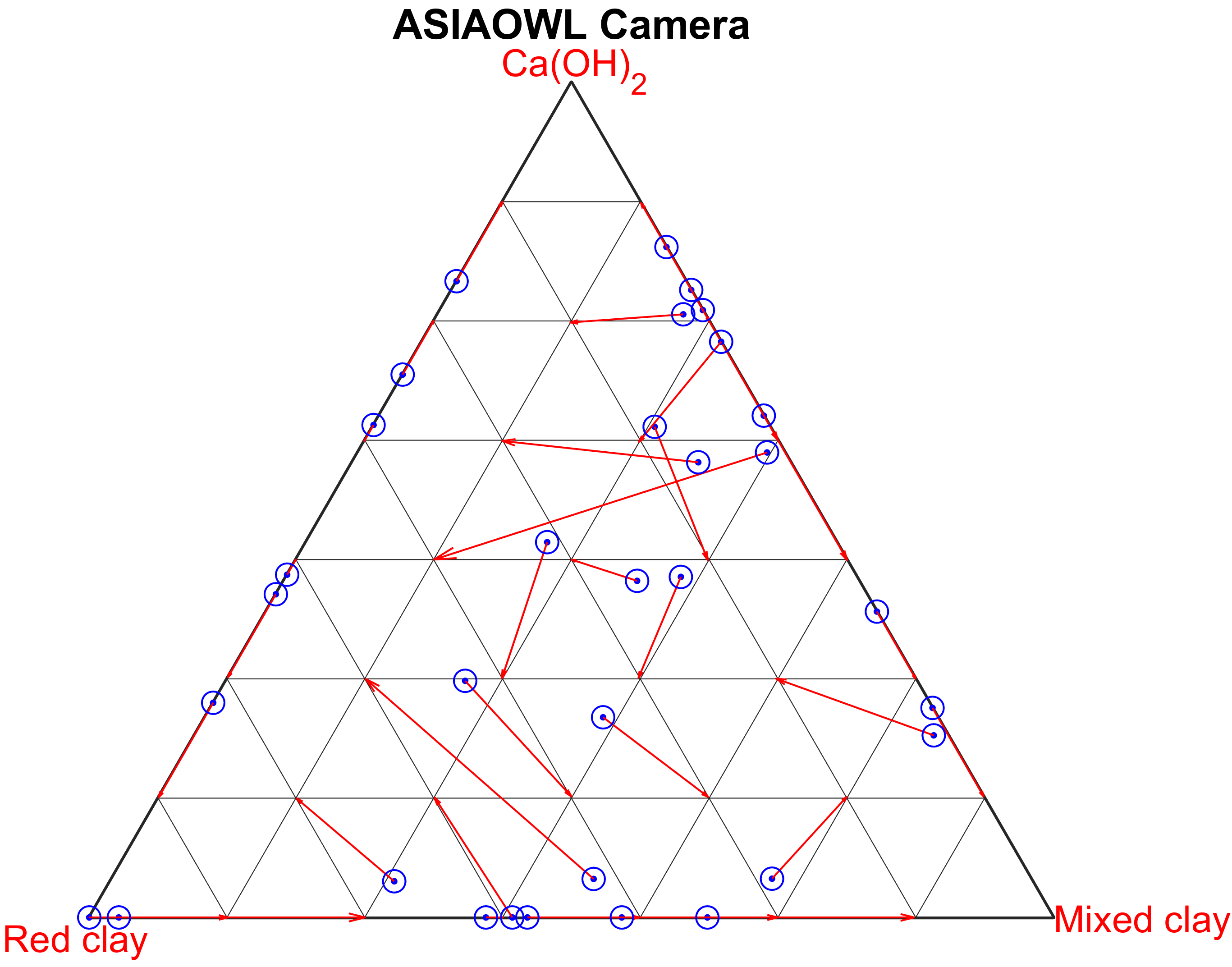}   \\
 (d) & (e) & (f) \\
  \end{tabular}
 \caption{{\color{black} Unmixing results overlaid on the ternary diagram of the mixtures of three clays; Top row: Linear unmixing; Bottom row: Hapke unmixing. In the figure, the blue dots denote the estimated positions of the mixtures while the red arrows show the real positions of the mixtures; (a and d) Agilent 4300 FTIR; (b and e) FX50 camera; (c and f) AisaOWL camera.}}
\label{Unmixing_results_MWIR}
\end{figure*}

Fig. \ref{Unmixing_results_MWIR} displays the estimated fractional abundances by FCLSU (Top row) and the Hapke model (Bottom row) on the reflectance dataset in the MWIR/LWIR, obtained from the binary and ternary mixtures of the Red Clay, Mixed Clay, and Calcium hydroxide, overlaid on the ternary diagram of the clay mixtures. As can be observed, the estimated fractional abundances deviate a lot from the true fractional abundances. From Fig. \ref{Unmixing_results_MWIR}, one can observe that the unmixing result is sensor and wavelength-dependent. {\color{black} FCLSU projected most of the ternary mixtures acquired by the Agilent 4300 FTIR dataset onto the line connecting mixed clay and Ca(OH)$_2$ while the Hapke model underestimated the fractional abundances of Mixed clay. On the other hand, both FCLSU and the Hapke model underestimated the fractional abundance of mixed clay in the FX50 dataset. In contrast to datasets acquired by Agilent 4300 FTIR and FX50, both FCLSU and the Hapke model underestimated the fractional abundance of Ca(OH)$_2$ in the AisaOWL dataset. }

\section{Data acquisition with X-Ray sensors}
\label{X-ray sensors}

It is well-known that in clay samples, the information depth (the distance below the surface from which information is contributing to the spectral reflectance) is limited in the optical wavelengths (400 nm to 2500 nm) (\cite{SADEGHI201566}). Although our samples were more than 1 cm thick, the information depth varies between a few micrometers and a few 100 micrometers, depending on the chemical composition and the porosity of the clay samples. The information contained in the reflectance matches with the ground-truth mass ratio only when the sample is sufficiently homogeneous. To verify if our samples are sufficiently homogeneous, we performed X-ray fluorescence (XRF) atomic/elemental analysis. X-ray fluorescence (XRF) is an analytical technique traditionally employed in geological sciences for qualitative and quantitative assessment of chemical elements in mineral samples \cite{Penner-Hahn2002}. The major reason to choose XRF analysis is that it is non-destructive and its information depth varies between a few micrometers and a few 100 micrometers (\cite{RAVANSARI2020105250}). However, to convert the mass ratios of our mixtures into atomic concentrations, the molecular composition of the pure clay samples is required. For this, X-ray powder diffraction is applied.

\subsection{X-ray powder diffractometer (XRPD)}
\begin{figure}[htb!]
\centering
 \begin{tabular}{l}
 	\includegraphics[width=.45\textwidth]{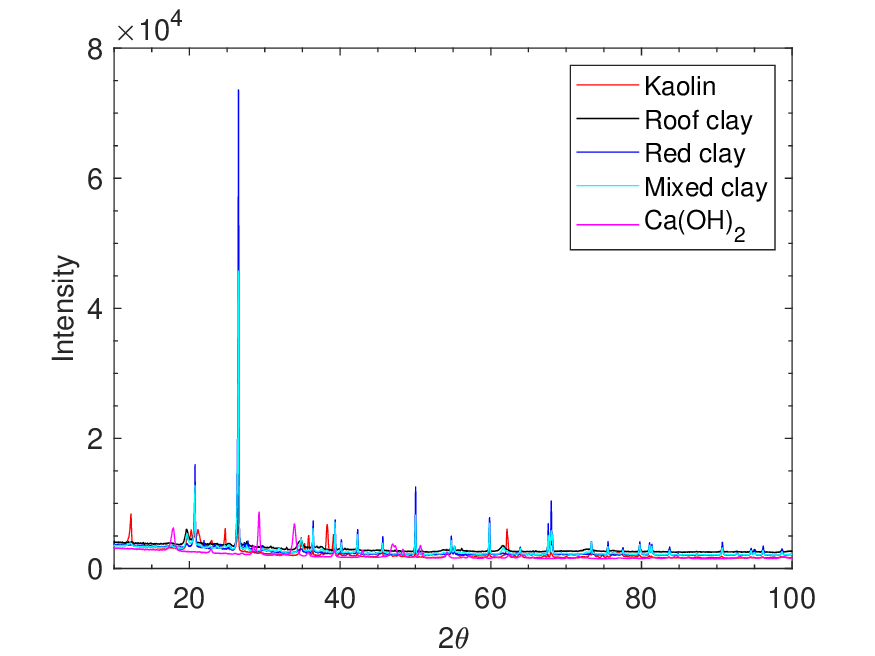}\\
   \end{tabular}
 \caption{The X-ray powder diffraction (XRPD) patterns obtained with a Huber G670 Guinier diffractometer.}
\label{HighqualityXRD}
\end{figure}

In this work, we utilized a Huber G670 Guinier diffractometer (Cu K$\alpha$1 radiation (X-ray wavelength 1.5406 Å), curved Ge(111) monochromator, transmission mode, image plate) to acquire the diffraction spectra of our five pure clay samples (see Fig. \ref{HighqualityXRD}). {\color{black} For the measurement of the diffraction pattern, the samples were dispersed in ethanol and spread over a polymeric transparent film. Guinier G670 Data Acquisition Program was used to read out the diffraction pattern generated on the imaging plate during X-Ray exposure. The program enabled saving the data in intensity vs. 2$\theta$ format for further treatment.} In the next step, {\color{black} Profex (version 5.2.0)} software (\cite{Doebelin2015ProfexAG}) was applied to determine the molecular compositions of the pure clay samples. The results are displayed in Table \ref{Pure sample summary}.

	\begin{table*}
		\caption{The molecular composition of the pure clay samples obtained by XRPD.}
		\centering
		\scalebox{0.85}{
		\begin{tabular}{|c|c|c|c|c|c|c|}
			\hline
			Molecule      			            & Molecular formula   & Kaolin  & Roof clay & Red clay & Mixed clay & Ca(OH)$_2$\\   
			\hline
   			Biotite  & Si$_{1.36}$ Al$_{1.24}$ Fe$_{1.4}$ Mg$_{0.71}$ Ti$_{0.16}$ Na$_{0.02}$ K$_{0.98}$ O$_{12}$ H$_{1.64}$  & 0\%  & 10.55\% & 3.76\% & 4.38\% & 0\%\\
                \hline
		    Calcium carbonate  & CaCO$_3$  & 0\%  & 0\% & 0\% & 0\% & 48.04\%\\
			\hline
			Calcium hydroxide  & Ca(OH)$_2$  & 0\%  & 0\% & 0\% & 0\% & 51.96\%\\
			\hline
			Goethite  & FeO(OH)  & 0\%  & 1.91\% & 0.5\% & 0.27\% & 0\%\\
			\hline
			Kaolinite  & Al$_2$Si$_2$O$_5$(OH)$_4$  & 87.24\%  & 45.32\% & 5.62\% & 39\% & 0\%\\
			\hline
			Silicon dioxide  & SiO$_2$  & 12.76\%  & 42.23\% & 90.13\% & 56.35\% & 0\%\\
			\hline
		\end{tabular}    }
		\label{Pure sample summary}
	\end{table*}
	

We then converted the ground truth mass ratios of the pure clay powders onto ground truth atomic concentrations.
{\color{black} After estimating the molecular concentrations for the pure clay powders, we proceeded to use XRF spectroscopy to estimate the elemental atomic concentrations of both pure clay powders and their mixtures.}

\subsection{micro-X-ray fluorescence spectrometer}
\begin{figure}[htb!]
\centering
 \begin{tabular}{l}
 	\includegraphics[width=.45\textwidth]{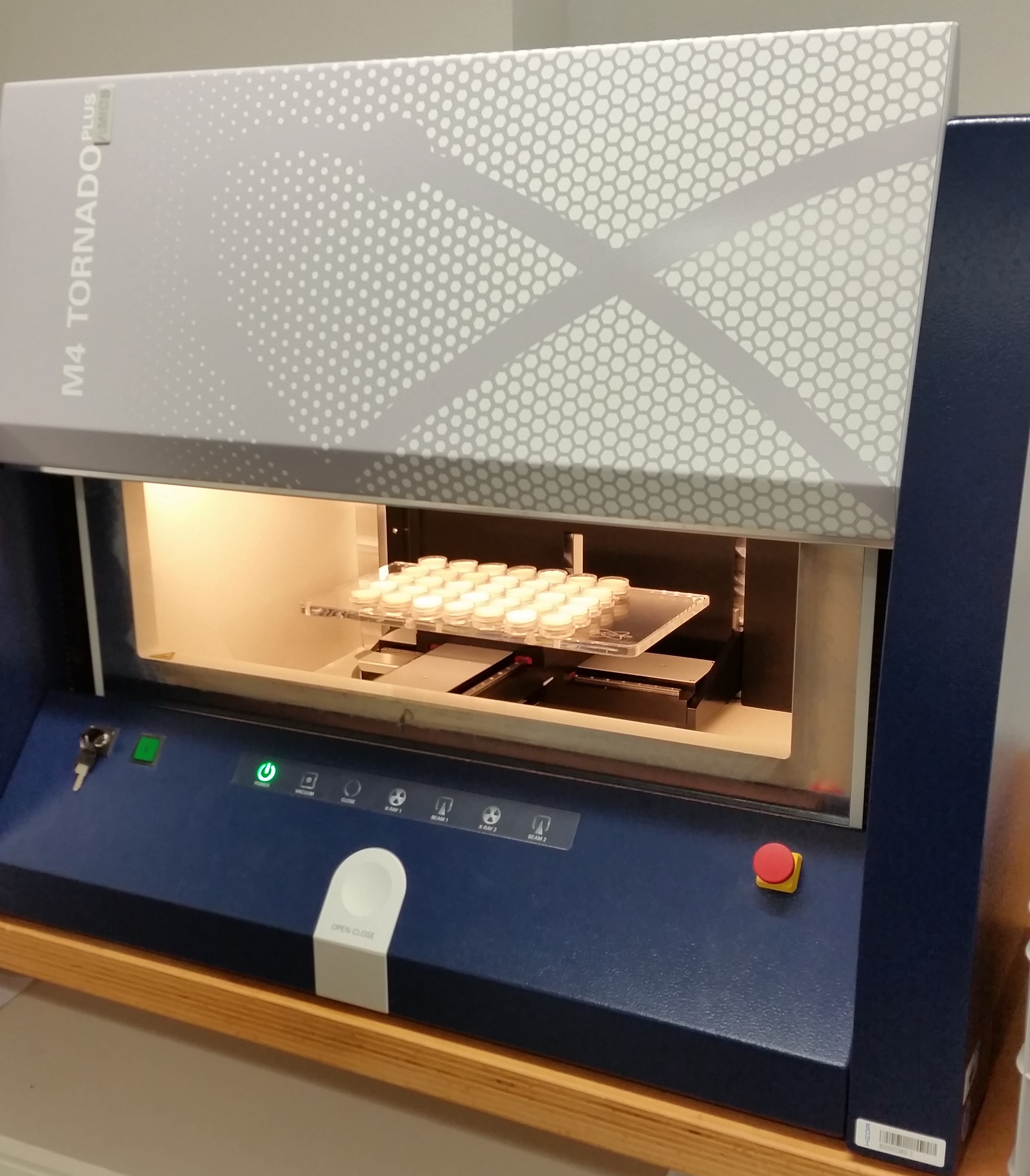}\\
   \end{tabular}
 \caption{The micro-X-ray fluorescence spectrometer (Bruker Tornado M4+).}
\label{XRF_Tornado}
\end{figure}

In this work, we employed a Bruker Tornado M4+ micro XRF spectrometer (see Fig. \ref{XRF_Tornado}) for elemental identification and quantification. Recent advances in the optical field (e.g. polycapillary lenses) allowed for the development of XRF spectrometers with high spatial resolution, referred to as micro-XRF ($\mu$XRF). The ability to focus on such small areas is particularly interesting for this study, as we aim to determine the elemental composition of powders with grain sizes of less than 200 $\mu$m.

\begin{figure}[htb!]
\centering
 \begin{tabular}{l}
	 \includegraphics[width=.45\textwidth]{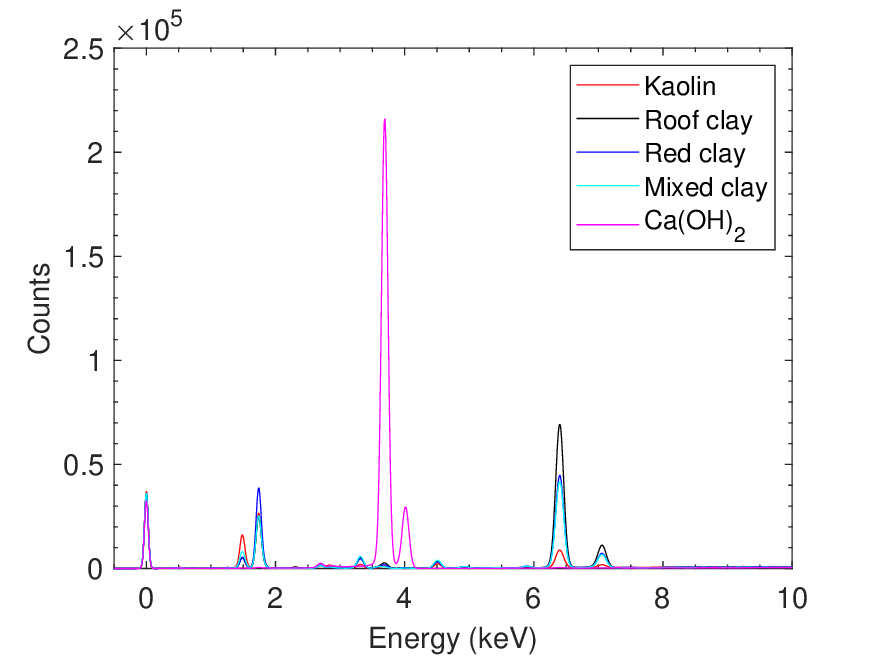}  \\
   \end{tabular}
 \caption{Spectra of the pure clay powders acquired by the micro-X-ray fluorescence spectrometer.}
\label{XRF spectra}
\end{figure}

Signal acquisition parameters were set as follows: a Rhodium anode is used as an excitation source (50 kV, 600 $\mu$A = 30 W of total power), collimated by polycapillary lenses and focused on the sample’s surface (spot size: 170 $\mu$m). X-ray signals emitted by our sample were collected for 15 seconds by two silicon detectors with Beryllium windows (XFlash\textregistered technology). The air pressure inside the analytical chamber was reduced to 2 mBar, aiming to reduce signal attenuation by atmospheric interactions and to increase detection sensitivity for lighter elements. We performed 3-point acquisitions on each sample to ensure the reproducibility and consistency of results. Fig. \ref{XRF spectra} shows the mean spectra of the pure clay powders acquired by the $\mu$XRF spectrometer. {\color{black} Elemental identification and quantification were performed by fundamental parameters algorithms derived from the Sherman equation \cite{SHERMAN1955283}.} 

While analyzing the results, we observed a sensor-specific bias in the atomic concentrations of the pure clays estimated by $\mu$XRF, when compared to the results from XRPD. To solve this issue, calibration is required. For this, we reconstruct the atomic concentrations of the clay powder mixtures estimated by $\mu$XRF as linear combinations of the atomic concentrations of the pure clay powders, to obtain the relative contributions of the pure clays in the mixtures (how much each pure clay contributes to the atomic concentration of the mixtures). This reconstruction comes down to applying FCLSU to the atomic concentrations of the mixtures. By multiplying these relative contributions with the ground truth atomic concentrations of the pure clay powders obtained by XRPD, bias-free atomic concentrations of the clay powder mixtures are obtained.

\begin{figure*}[htbp]
\centering
 \begin{tabular}{cc}
 	\includegraphics[width=.45\textwidth]{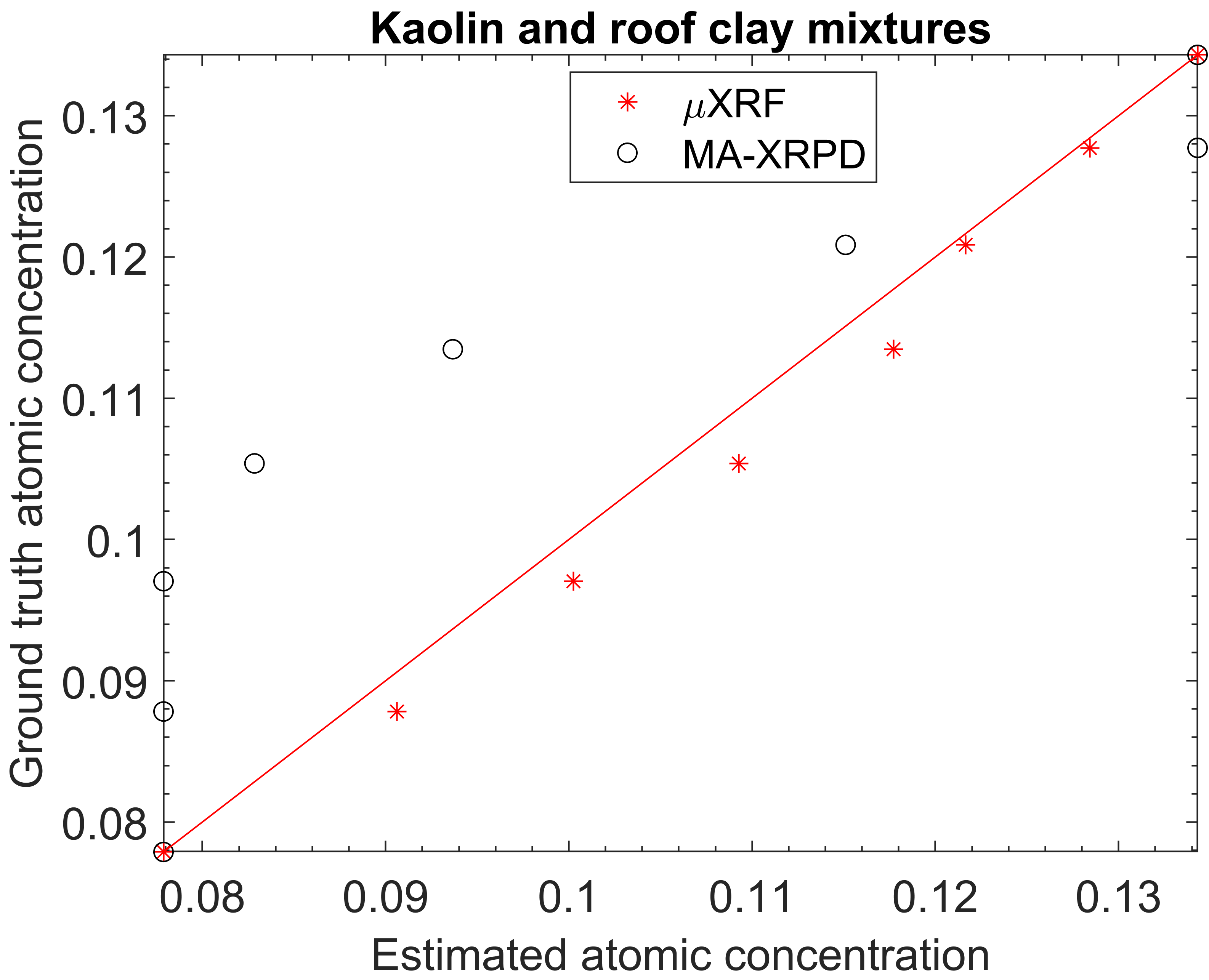}& \includegraphics[width=.45\textwidth]{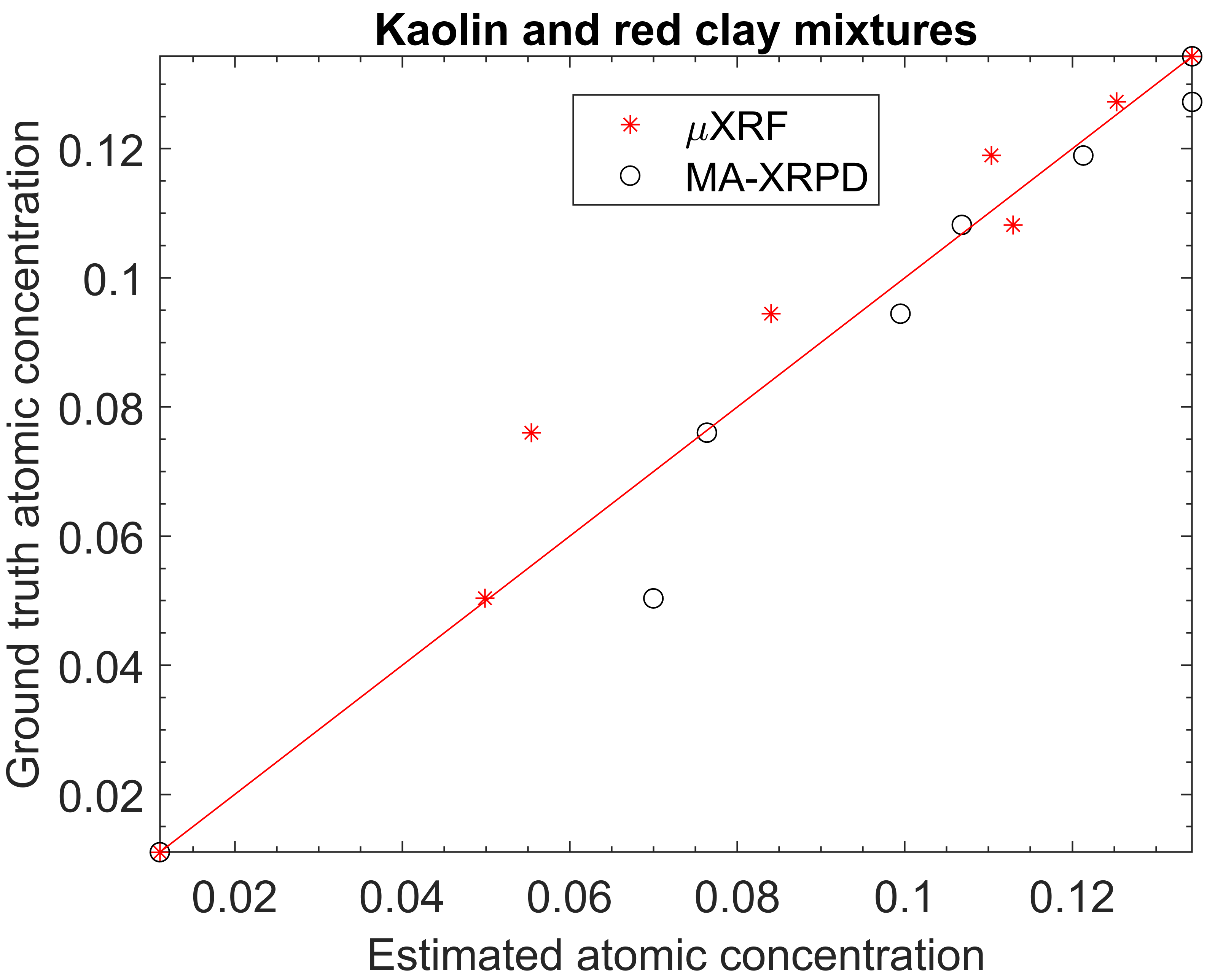}\\
 (a) & (b) \\
 \includegraphics[width=.45\textwidth]{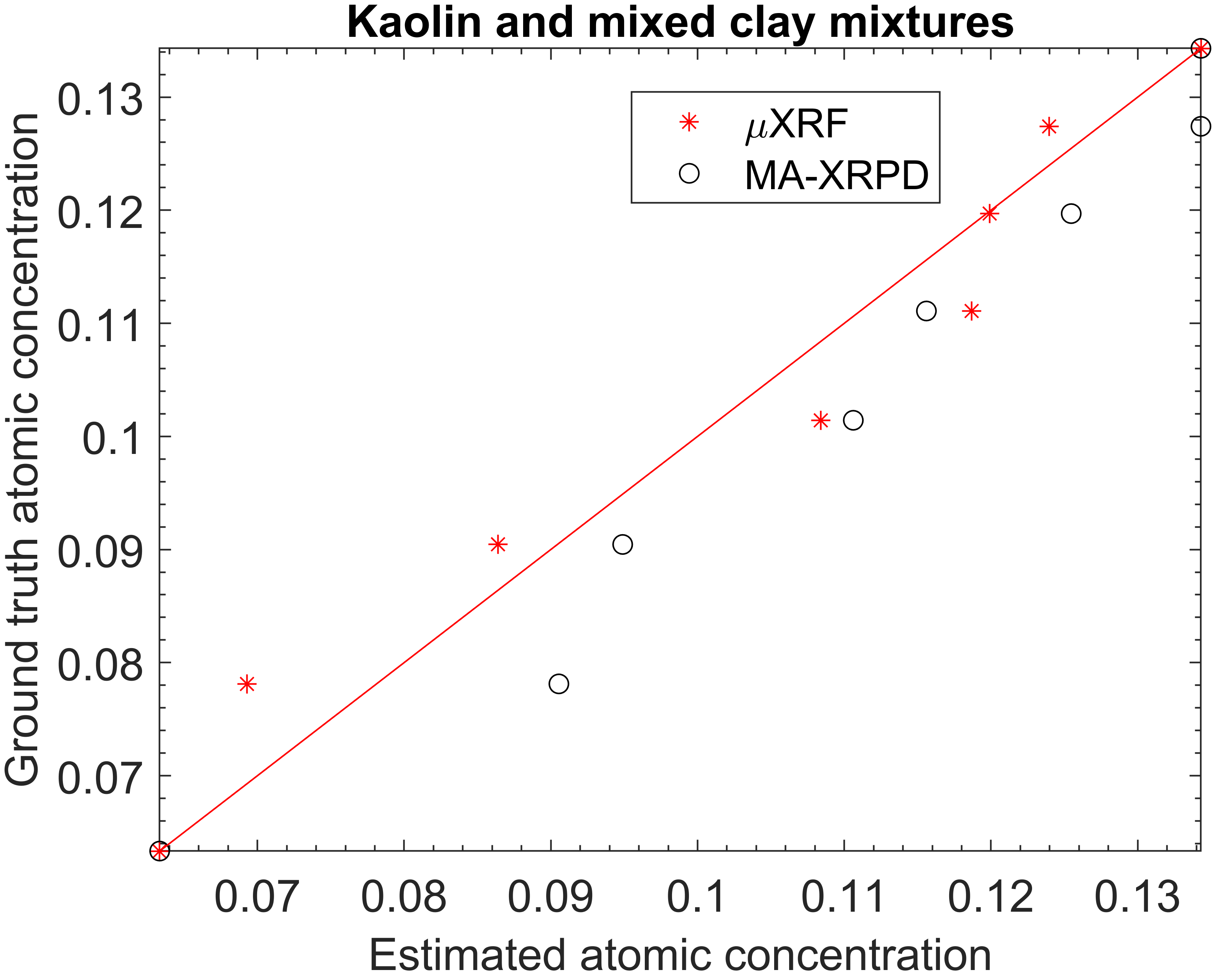} &
 \includegraphics[width=.45\textwidth]{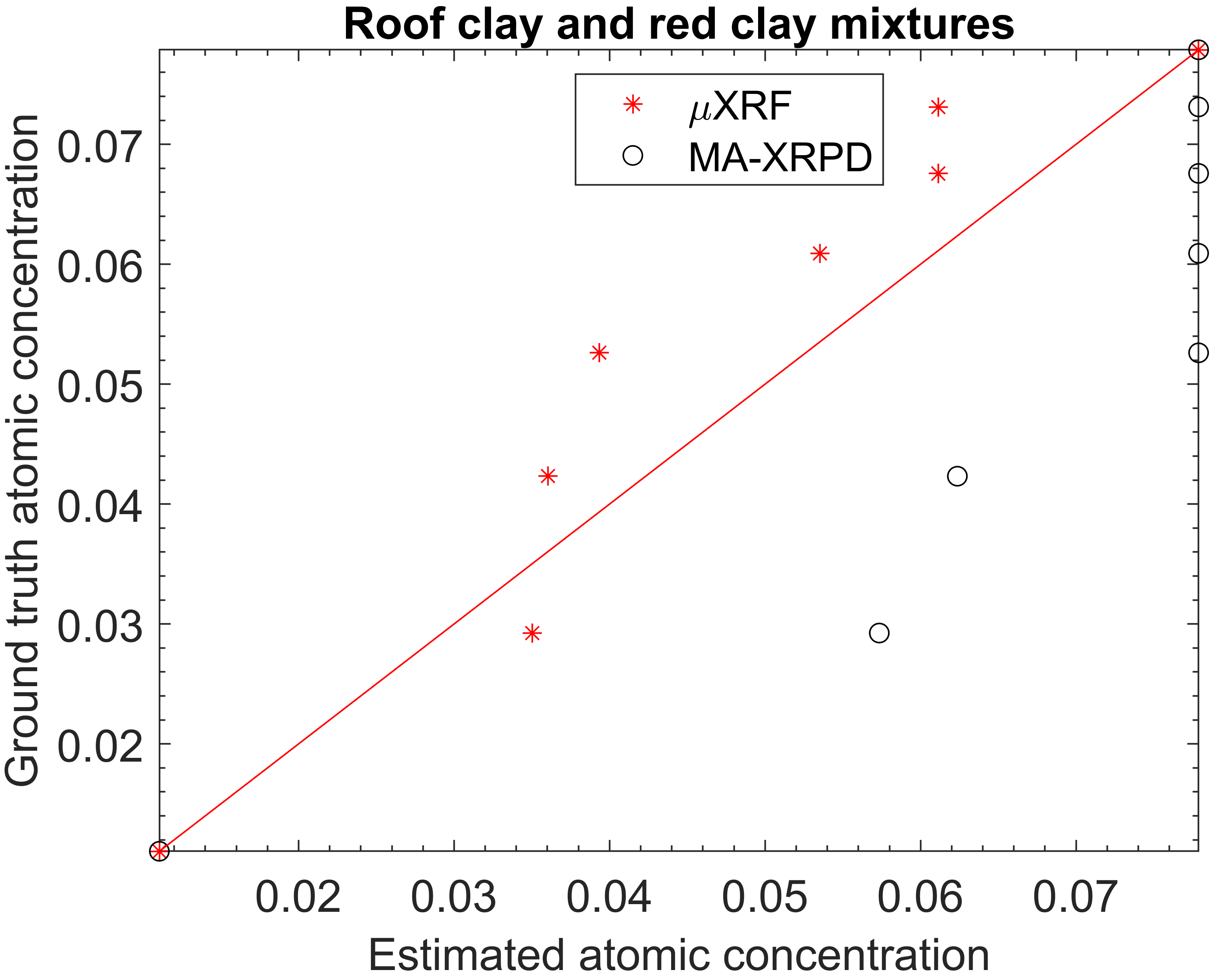}\\
 (c) & (d) \\
  \end{tabular}
 \caption{Atomic concentration of Aluminium estimated by calibrated $\mu$XRF and MA-XRPD versus ground truth; (a) Kaolin and roof clay mixtures; (b) Kaolin and red clay mixtures; (c) Kaolin and mixed clay mixtures; (d) Roof clay and red clay mixtures.}
\label{XRF_vs_GT}
\end{figure*}

In Fig. \ref{XRF_vs_GT}, a scatterplot displays the atomic concentration of Aluminum, estimated by the calibrated $\mu$XRF versus the ground truth, on several binary mixtures. As can be observed, the atomic concentration estimated by $\mu$XRF perfectly matches the ground truth atomic concentrations. This demonstrates that our samples were quite homogeneous.

\subsection{ Macro X-ray powder diffractometer}

\begin{figure}[htbp]
\centering
 \begin{tabular}{cc}
 	\includegraphics[width=.15\textwidth]{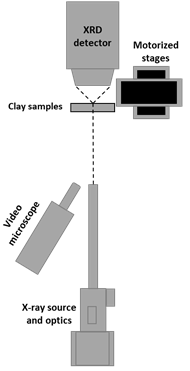}& 		\includegraphics[width=.24\textwidth]{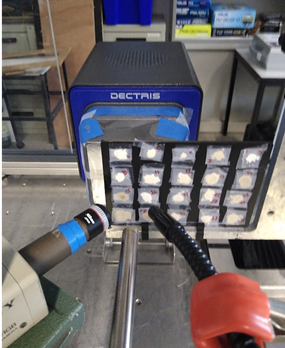}\\
 (a) & (b) \\
   \end{tabular}
 \caption{(a) Schematic; (b) a photograph of the MA-XRPD instrument in transmission geometry.}
\label{X-ray_diffractometer}
\end{figure}

Next to $\mu$XRF, we employed a macroscopic X-ray powder diffractometer (MA-XRPD) to estimate the atomic concentrations in all 330 samples. The MA-XRPD instrument employs a low-power Cu-anode X-ray micro source (30 W, I$\mu$S–Cu, Incoatec GmbH, DE) that delivers a monochromatic (Cu–K$\alpha$) and focused X-ray beam (focal spot diameter, 400 $\mu$m; output focal distance, 39.8 cm; divergence, 2.6 mrad). A PILATUS 200 K detector placed perpendicular to the source at the output focal distance collects diffraction patterns in transmission mode. The samples were positioned in front of the area detector at a distance of 3 cm (Fig. \ref{X-ray_diffractometer}). Because samples had to be held vertically to measure transmitted radiation, a thin layer of a clay sample was prepared and attached to a 3D-printed sample holder with an opening of 10$\times$ 15 mm. XY motorized stages (maximum travel ranges, 10 cm × 25 cm; Newport Corporation, Irvine, CA) allow for the movement of the samples during the imaging experiment while the instrument remains stationary. The samples are positioned by use of a video camera with a limited depth of field. The entire scanning operation, including motor movements and acquisitions, is controlled by in-house software. A LaB6 standard for powder diffraction (SRM 660, NIST) is used for the calibration of the instrument. The analytical characteristics of the MA-XRPD system have recently been reported elsewhere (\cite{Vanmeert2018}). {\color{black} From each sample, a minimum of six measurements were taken.}

\begin{figure}[htb!]
\centering
 \begin{tabular}{l}
 	\includegraphics[width=.45\textwidth]{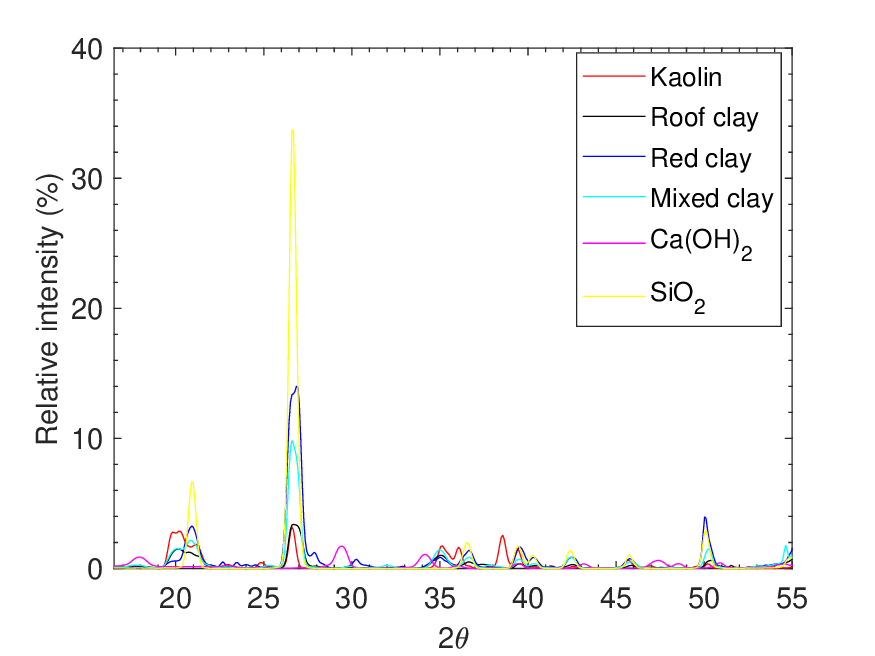} \\
   \end{tabular}
 \caption{Spectra of the pure clay powders acquired by the macroscopic X-ray powder diffractometer.}
\label{Xray spectra}
\end{figure}

{\color{black} The 2D diffraction patterns measured by MA-XRPD were converted to 1D diffraction patterns by utilizing the XRDUA software package \cite{DeNolf:po5003}. Fig. \ref{Xray spectra} shows the mean 1D diffraction patterns of the pure clay powders. Because most clay powders contain Silicon dioxide (see peak around 26.5 2$\theta$), its diffraction spectrum is shown as well. The diffraction patterns of the pure clay powders and the clay mixtures were analyzed by using the {\color{black} Profex (version 5.2.0)} software to estimate atomic concentrations.} Similar to $\mu$XRF, we corrected for sensor-specific bias to obtain bias-free atomic concentrations. Fig. \ref{XRF_vs_GT} indicates that results of MA-XRPD are far less accurate than $\mu$XRF at predicting the atomic concentrations of the pure and mixed clay powders. {\color{black} This could be explained by the fact that MA-XRPD only analyses the crystalline phases while $\mu$XRF records the atomic concentrations of both the crystalline and amorphous phases}.

\section{Discussion}
\label{Discussion}
From the experiments, the following general conclusions can be drawn:
\begin{itemize}
    \item The reflectance spectra of intimate mineral powder mixtures are highly dependent on the acquisition conditions, i.e., variable illumination conditions, distance, and orientation from the sensor, the use of different sensors, and white calibration panels. {\color{black} In general, band-wise scaling differences (see Fig. \ref{Mineralsspectra}) can be observed between the spectral reflectances acquired by the different sensors.}

    \item The five clay powders studied in this work have spectral features around 1400 nm, 1900 nm, and
    between 2100 and 2500 nm. These features indicate the presence of vibrational hydroxyl processes. Although the spectra of roof clay, red clay, and mixed clay showed similarities, in general, the overall spectral shape and reflectance values of the pure clay powders are distinctive (see Fig. \ref{Mineralsspectra}). This is a prerequisite for differentiating these five clay powders in the mixtures.
  
    \item {\color{black} Both the linear mixing model and the Hapke model are found to be not suitable for predicting the composition of clay powder mixtures accurately in the VNIR/SWIR wavelength regions. The linear mixing model projected most of the ternary mixtures onto the faces of the simplex leading to a significant error in the estimated fractional abundances. To accurately predict the fractional abundances of intimate mixtures, advanced nonlinear unmixing algorithms are required that can tackle both nonlinearity and spectral variability.}

    \item In MWIR/LWIR wavelength regions, the studied clay powders have multiple absorption features (2500 nm to 15385 nm) that can be related to the fundamental stretching and bending vibrations of their fundamental functional groups, i.e., the OH and Si-O group \cite{MADEJOVA2017107}. Besides an intensity difference, Kaolin, Roof clay, Red clay, and Mixed clay have the same spectral shape between 2500 nm and 6000 nm, while Ca(OH)2 shows a significantly different spectral shape. The feature between 8000 nm and 10000 nm indicates the presence of the Si-O group in roof clay, red clay, and mixed clay.

    
    \item {\color{black} The unmixing results on the dataset acquired in the MWIR/LWIR are found to be sensor and wavelength-dependent (see Fig. \ref{Unmixing_results_MWIR}). The unmixing results are worse than those estimated in the VNIR/SWIR wavelength regions.}
    
    \item {\color{black} Although our clay powders have specific absorption features in the MWIR and LWIR, it is not clear to what extent the MWIR and LWIR can contribute to the estimation of the composition of intimate mixtures. This has to be investigated in detail.}

    \item Because the information depth of the optical wavelengths (400 nm to 2500 nm) is limited in clay samples, the bulk composition of the mixture matches the information contained in the reflectance dataset, only when the sample is sufficiently homogeneous. The high correlation between the atomic concentration estimated by $\mu$XRF and the ground truth (bulk) atomic concentrations demonstrates that our samples were homogeneous.  

    \item {\color{black} Although we acquired datasets by 13 different sensors, the reflectance dataset acquired by the following five sensors has to be adequately preprocessed before applying them for spectral unmixing: a) Senops HSC2 reflectance spectra are entirely different from those acquired by other sensors in the VNIR wavelength region (see Fig. \ref{Mineralsspectra}). This might be due to inaccurate radiometric calibration; b) Approximately 40 \% of the spectral reflectance dataset acquired by the Cubert Ultris X20P and the Cubert panchromatic camera are over-saturated. This  may be  due to the  workflow to convert the raw dataset to reflectance, and it has to be investigated in more detail; c) We could not accurately convert the raw Telops MWIR data into reflectance. In our opinion, the measurement is inaccurate, so we will only upload the raw data; d) The dataset acquired by the Telops Hypercam contains a significant amount of emission information overlaid onto diffuse reflection. To remove emission information, a physical model has to be developed that can describe the measured data as a mixture of diffuse reflection and emission information; e) Although the dataset acquired by Specim AisaFenix for most of the samples is accurate, the spectral reflectance between 1500 nm and 1900 nm show unnatural behavior (see Fig. \ref{Mineralsspectra}). This unnatural behavior is much more visible in the spectral reflectance of Ca(OH)$_2$ (see Fig. \ref{Mineralsspectra} (e)). We could not find a proper explanation for this distortion.      
    }

    
\end{itemize}

\section{Conclusions}
\label{sec:conclusion}

In this work, we generated 325 samples by homogeneously mixing five different clay powders (Kaolin, Roof
clay, Red clay, mixed clay, and Calcium hydroxide). Among the 325 samples, 60 mixtures were binary, 150 were ternary, 100 were quaternary, and 15 were quinary. These samples (and pure clay powders) were scanned by 13 different sensors, with a wavelength range between the visible and the long-wavelength infrared regions (i.e., between 350 nm and 15385 nm) to produce a comprehensive hyperspectral dataset of intimate mixtures. We verified that the generated samples were sufficiently homogeneous by performing X-ray powder diffraction and X-ray fluorescence elemental analysis. The low performance of the linear mixing model to estimate the composition of the mixtures demonstrates that advanced hyperspectral unmixing methods are required that can tackle both spectral variability and nonlinearity.

\section*{Acknowledgment} 
The research presented in this paper is funded by the Research Foundation-Flanders - project G031921N. The authors would like to thank the research group Electron Microscopy for Materials Science (EMAT) for the use of their Huber G670 Guinier diffractometer. The authors would like to thank Dr. Maria Batuk for helping us to analyze our pure clay powders. PSR 3500, Agilent 4300 FTIR, Telops Hypercam LWIR, Specim AisaFENIX: The Helmholtz Institute Freiberg for Resource Technology is gratefully thanked for supporting and funding this project. Telops MWIR, Specim sCMOS: This research was supported by EIT RawMaterials GmbH within the KAVA up-scaling project “inSPECtor”. Specim AisaOWL: Funded by the European Regional Development Fund and the Land of Saxony. Senops HSC2: This sensor was acquired within the scope of the NEXT project, which has received funding from the European Union’s Horizon 2020 research and innovation programme under Grant Agreement No. 776804 – H2020–SC5–2017. Cubert Ultris X20P: Funded by the Federal Ministry of Education and Research of Germany in the framework of the Competence Cluster Recycling and Green Battery (greenBatt) under BMBF-No. 03XP0337, within the scope of the DIGISORT project.
Specim JAI, Specim FX50: Acquired during the HighSpeedImaging project, Funded by the European Regional Development Fund and the Land of Saxony. Bruker Tornado M4+: The Bruker TORNADO M4plus was financed by the Sächsische Aufbau Bank (SAB), via the Highspeed Images fonds, No. 5722280116.

\bibliographystyle{IEEEbib}
\bibliography{refs}
\end{document}